\documentclass[10pt,twocolumn,letterpaper]{article}

\usepackage{cvpr}              % To produce the CAMERA-READY version
\usepackage[accsupp]{axessibility}  % Improves PDF readability for those with disabilities.
\usepackage[table,dvipsnames]{xcolor}
\usepackage{color,colortbl}
\usepackage{multirow}
\usepackage{array}
\usepackage{float}

\usepackage{subcaption}
\usepackage[font=small,labelfont=bf]{caption}

\usepackage{graphicx}
\usepackage{amsmath}
\usepackage{amssymb}
\usepackage{arydshln}

\usepackage[export]{adjustbox}

\usepackage{soul}

\definecolor{LightGray}{rgb}{0.95, 0.95, 0.95}
\definecolor{LightBlue}{rgb}{0.89, 0.93, 0.99}
\definecolor{LightRed}{rgb}{0.99, 0.93, 0.89}

\usepackage[pagebackref,breaklinks,colorlinks]{hyperref}

\usepackage[capitalize]{cleveref}
\crefname{section}{Sec.}{Secs.}
\Crefname{section}{Section}{Sections}
\Crefname{table}{Table}{Tables}
\crefname{table}{Tab.}{Tabs.}

\def\cvprPaperID{7000} % *** Enter the CVPR Paper ID here
\def\confName{CVPR}
\def\confYear{2022}

\begin{document}

\title{DiffusioNeRF: Regularizing Neural Radiance Fields \\ with Denoising Diffusion Models
}
\author{
\hspace{10pt} Jamie Wynn \hspace{10pt} Daniyar Turmukhambetov
\\
Niantic\\
\href{https://www.github.com/nianticlabs/diffusionerf}{www.github.com/nianticlabs/diffusionerf}
}

\maketitle

%%%%%%%%% ABSTRACT
\begin{abstract}
   Under good conditions, Neural Radiance Fields (NeRFs) have shown impressive results on novel view synthesis tasks. NeRFs learn a scene's color and density fields by minimizing the photometric discrepancy between training views and differentiable renderings of the scene. Once trained from a sufficient set of views, NeRFs can generate novel views from arbitrary camera positions. However, the scene geometry and color fields are severely under-constrained, which can lead to artifacts, especially when trained with few input views.
   
   To alleviate this problem we learn a prior over scene geometry and color, using a denoising diffusion model (DDM). Our DDM is trained on RGBD patches of the synthetic Hypersim dataset and can be used to predict the gradient of the logarithm of a joint probability distribution of color and depth patches. We show that, these gradients of logarithms of RGBD patch priors serve to regularize geometry and color of a scene. During NeRF training, random RGBD patches are rendered and the estimated gradient of the log-likelihood is backpropagated to the color and density fields. Evaluations on LLFF, the most relevant dataset, show that our learned prior achieves improved quality in the reconstructed geometry and improved generalization to novel views. Evaluations on DTU show improved reconstruction quality among NeRF methods.
   
\end{abstract}

%%%%%%%%% BODY TEXT
\section{Introduction}\label{sec:intro}

\begin{figure}[ht]
\centering
\begin{minipage}[c]{\linewidth}
\begin{minipage}[c]{\linewidth}
\centering
  \begin{minipage}[c]{0.48\linewidth}
  \includegraphics[width=0.99\linewidth]{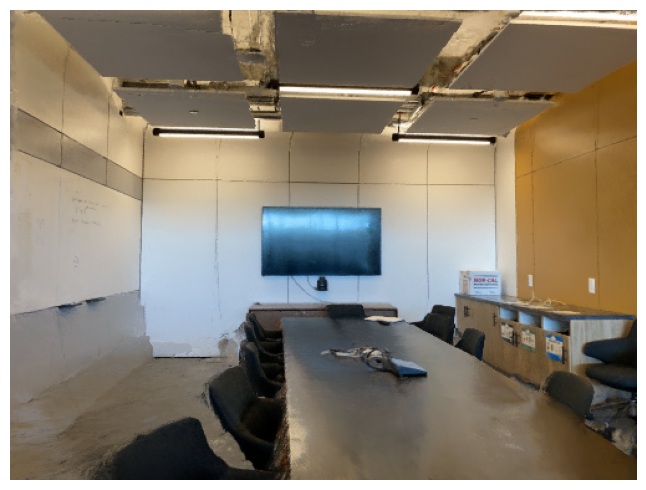}
  \end{minipage}
  \begin{minipage}[c]{0.48\linewidth}
  \includegraphics[width=0.99\linewidth]{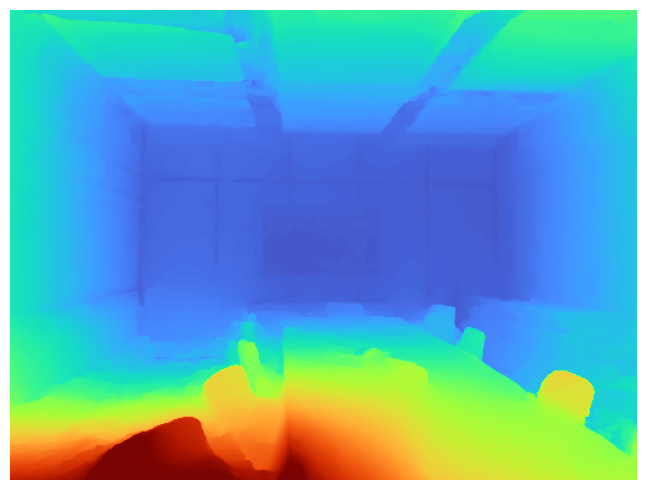}
  \end{minipage}
\end{minipage}
\\
\begin{minipage}[c]{\linewidth}
\centering
\small
\vspace{1pt}
(a) DiffusioNeRF (Ours)
\vspace{1pt}
\end{minipage}
\\
\begin{minipage}[c]{\linewidth}
\centering
  \begin{minipage}[c]{0.48\linewidth}
  \includegraphics[width=0.99\linewidth]{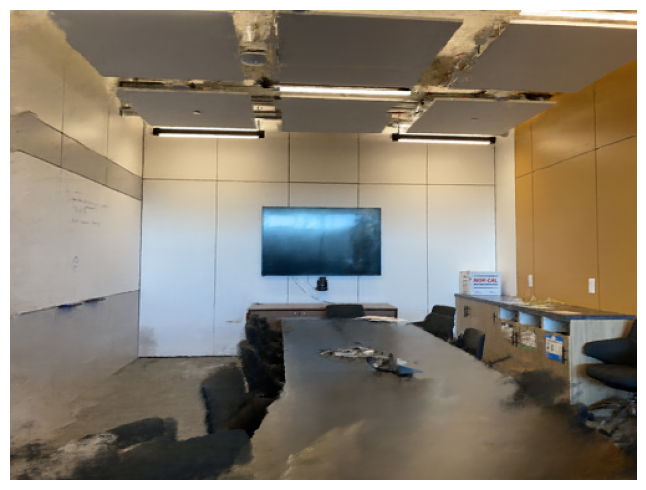}
  \end{minipage}
  \begin{minipage}[c]{0.48\linewidth}
  \includegraphics[width=0.99\linewidth]{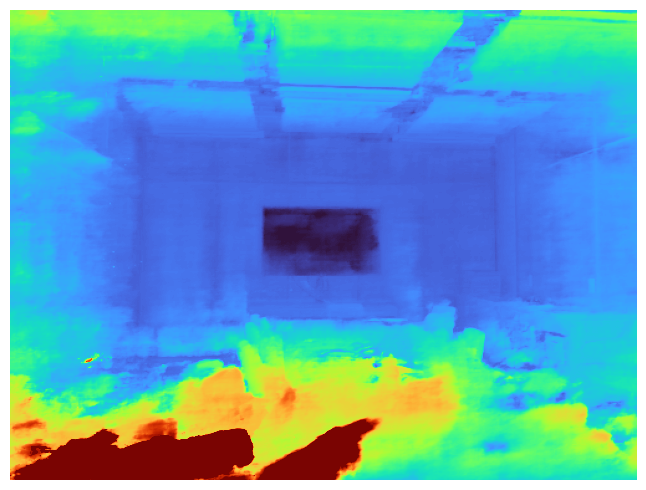}
  \end{minipage}
\end{minipage}
\\
\begin{minipage}[c]{\linewidth}
\centering
\small
\vspace{1pt}
(b) RegNeRF~\cite{Niemeyer2021Regnerf}
\end{minipage}
% \\
% \begin{minipage}[c]{\linewidth}
% \centering
%   \begin{minipage}[c]{0.48\linewidth}
%   \includegraphics[width=0.99\linewidth]{figures/teaser/mipnerf_001_rgb.png}
%   \end{minipage}
%   \begin{minipage}[c]{0.48\linewidth}
%   \includegraphics[width=0.99\linewidth]{figures/teaser/mipnerf_001_depth.png}
%   \end{minipage}
% \end{minipage}
% \\
% \begin{minipage}[c]{\linewidth}
% \centering
% \small
% \vspace{1pt}
% (c) Mip-NeRF 360~\cite{barron2022mipnerf360}
% \vspace{1pt}
% \end{minipage}
\end{minipage}
\caption{Image and depth map rendered from a test view. All NeRF models were trained with \textbf{3} views of the LLFF~\cite{mildenhall2019llff} dataset's ``Room'' scene. Our priors encourage NeRF to explain the TV and table geometry with flat surfaces in the density field, and to explain the view-dependent color changes with the color field.}
\label{fig:teaser}
\end{figure}

Neural radiance fields, neural implicit surfaces, and coordinate-based scene representations are proving valuable for novel view synthesis and 3D reconstruction tasks. NeRFs~\cite{mildenhall2020nerf} learn a \emph{specific scene's} appearance as a multi-layer perceptron that predicts density and color, when given any 3D point and a viewing direction.

This volume representation allows differentiable rendering from arbitrary views, where predicted color contributions along a ray are alpha-composited according to the density predictions.

The model is trained with the aim of faithfully reconstructing images captured with known camera poses.
Even when trained with just a photometric reconstruction loss, NeRFs show impressive generalization capabilities, inspiring novel applications in virtual and augmented reality, and visual special effects.

However, with small numbers or even with large numbers of input views, the scene color and geometry fields are severely under-constrained. Indeed, an infinite number of NeRFs can explain all training views. In practice, NeRFs can generate low-quality and physically implausible geometries and surface appearances. For example, ``floaters'' are one common artifact, where the fitted density field contains clouds of semi-transparent material floating in mid-air that would look reasonable in 2D once rendered from training views, but look implausible from novel views.

Various hand-crafted regularizers and learned priors have been proposed to tackle these issues: hand-engineered priors to constrain the scene geometry~\cite{barron2022mipnerf360,Niemeyer2021Regnerf}, learned priors that force plausible renderings from arbitrary views~\cite{Niemeyer2021Regnerf}, and methods that use single image depth and normal estimation~\cite{Yu2022MonoSDF,wang2022neuris} to provide high-level constraints on the estimated scene geometry. However, there are no approaches that learn a joint probability distribution of the scene geometry and color.

Our contribution is leveraging denoising diffusion models (DDMs) as a learned prior over color and geometry. Specifically, we use an existing synthetic dataset to generate a dataset of RGBD patches to train our DDM.
DDMs do not predict a probability for RGBD patch distribution. Rather, they provide the gradient of the log-probability of the RGBD patch distribution, \ie the negative direction to the noise predicted by DDM is equivalent to moving towards the modes of the RGBD patch distribution.
As NeRFs are trained with stochastic gradient descent, gradients of log-probabilities are sufficient, as they can be backpropagated to NeRF networks during training to act as a regularizer; probabilities are not required for this purpose. We demonstrate that the DDM gradient encourages NeRFs to fit density and color fields that are more physically plausible on the LLFF and DTU datasets.

%------------------------------------------------------------------------
\section{Related work}\label{sec:related_work}

\noindent
\textbf{Geometry modeling}
The geometry of the scene can be modeled as a density field~\cite{mildenhall2020nerf}, occupancy field~\cite{DVR,Oechsle2021ICCV} or signed distance field~\cite{yariv2020multiview,yariv2021volume,wang2021neus}. Geometry models can be rendered using differentiable surface/volumetric rendering, so that the training loss for a NeRF model is the photometric reconstruction loss~\cite{mildenhall2020nerf}. Signed distance fields also require regularization with an Eikonal loss~\cite{eikonal} to constrain the distance field to be valid. Our regularizer operates on rendered color and depth patches, so it can be applied to any geometry representation.

\noindent
\textbf{Field representation}
NeRFs~\cite{mildenhall2020nerf} represent geometry with a multi-layer perceptron that is queried with a 3D coordinate. Positional encoding of coordinates, where coordinate values are evaluated with sinusoids at different frequencies, allows modeling of high-frequency density signals with MLPs~\cite{tancik2020fourfeat}.
Alternatively, \cite{yu_and_fridovichkeil2021plenoxels,hedman2021snerg} encode scalar opacity and spherical harmonic coefficients in a sparse voxel representation, and shows that novel views can be synthesized without MLPs.
Similarly, Neural Sparse Voxel Fields~\cite{liu2020neural} stores feature encodings in a sparse voxel octree structure that can be trilinearly interpolated and passed through an MLP to predict density and color, thus improving the modeling capacity and rendering speed of NeRFs. 
MVSNeRF~\cite{mvsnerf} predicts a volume of feature encodings by constructing a 3D cost volume and processing it with 3D CNNs. Density and color MLPs trilinearly interpolate the feature encoding volume to train NeRFs. The 3D CNN can be pretrained on a large number of scenes, which allows faster convergence on novel scenes.

Instant Neural Graphics Primitives~\cite{mueller2022instant} uses multi-scale hash tables to store feature encodings of all coordinates in a fixed memory block. This allows storing features at varying spatial resolutions, and consequently reduces the size of the MLP that models geometry and color. With a GPU-optimized implementation, Instant NGP can train NeRFs in minutes without quality degradation. Our contribution is in priors used for NeRF optimization, and hence our method is agnostic to the underlying geometry representation. As Instant NGP is fast to train and render, we use it as a backbone for our experiments.

\noindent
\textbf{Density regularization}
Mip-NeRF 360~\cite{barron2022mipnerf360} proposes a density regularizer that encourages compactness of the density along conical frustums. In addition to our learned regularizer, we use \cite{barron2022mipnerf360} density regularizer as it helps to sharpen the distribution of densities along sampled rays.

\noindent
\textbf{Regularization with loss terms}
Loss terms to regularize NeRFs can play an important role in the final result, as they provide additional supervision to under-constrained geometry and color fields. Some regularizers are hand-crafted to encourage depth and normal smoothness, \eg \cite{zhang2021nerfactor,Niemeyer2021Regnerf,Oechsle2021ICCV,barron2022mipnerf360}. In \cite{Jain_2021_ICCV}, a semantic loss is introduced to make high-level semantic attributes consistent across renderings from random views.
In~\cite{roessle2022depthpriorsnerf} a loss term regularizes rendered depth maps with depths estimated using Structure-from-Motion and depth completion methods. MonoSDF~\cite{Yu2022MonoSDF} regularizes occupancy fields with loss terms that incorporate depth and normals maps predicted with a single-image depth prediction model. Similarly, \cite{wang2022neuris} introduces loss terms that use a single-image normal prediction model to regularize rendered normal maps.  While all these approaches introduce high-level geometric supervision to NeRFs, the predicted depth and normals are fixed during NeRF fitting and hence the depth and normal models provide a unimodal prior over geometry. Furthermore, the additional supervision is not adapted to the NeRF reconstructions and hence the monocular depth and normal predictions are trusted blindly.

\begin{figure*}[th]
  \centering
   \includegraphics[width=0.94\linewidth]{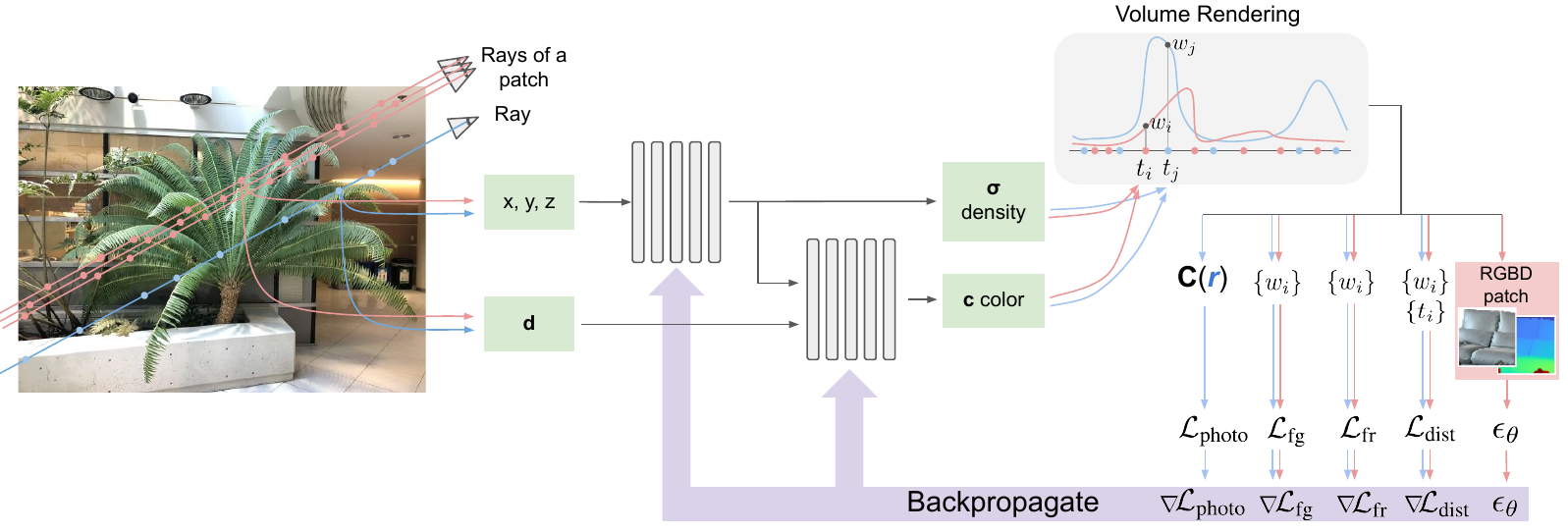}
   \caption{Illustration of our method. The scene is sampled with {\color{Cerulean} training-view rays} and {\color{Salmon} rays originating from random patches}. Color and density are predicted by MLP for the 3D points sampled along the rays. Volumetric rendering is used to estimate expected color $\mathbf{C}(\mathbf{r})$, depth $\mathbf{D}(\mathbf{r})$ as well as weights of color contributions $\{ w_i\}$ and positions of samples $\{ t_i \}$. These estimates are used to compute gradients of losses that are backpropagated to color and density MLPs. DDM model $\epsilon_\theta$ uses {\color{Salmon} RGBD patches} to predict color and density gradients that are passed to MLPs directly.
   Instant NGP's multi-scale hash table of feature encodings is not illustrated for simplicity.}
   \label{fig:pipeline}
\end{figure*}

\noindent
\textbf{Regularization with Normalizing Flows}
RegNeRF~\cite{Niemeyer2021Regnerf} uses a 2D depth patch smoothness prior and a normalizing flow model as a learned prior over 2D RGB patches. The color patches are rendered while fitting the NeRF and a term proportional to the log probability density assigned to the patch by the normalizing flow model is added to the loss function.

However, the underlying cause of NeRF's dramatic performance degradation in the few-view case is that the geometry is poor, so we argue that it is preferable to regularize the geometry directly, rather than indirectly via RGB patches. By learning a distribution over RGBD patches we also benefit from the fact that color and depth are strongly correlated, and therefore attempting to regularize them separately discards information.

RegNeRF~\cite{Niemeyer2021Regnerf} uses MLPs to model color and density fields, hence during NeRF training the patch rendering cost can extend NeRF training time substantially. Thus, RegNeRF renders $8 \times 8$ patches for the prior model, which severely limits the amount of context visible to the normalizing flow model. We use Instant NGP for our NeRF representation, which has a fast rendering time, allowing us to model priors over $48 \times 48$ patches.

Normalizing flows are generative models that learn to transform a simple probability distribution into a more complex data distribution~\cite{kobyzev2020normalizing}. The model is built of blocks that fulfil the requirements of (i) preserving the number of dimensions of input and output features; (ii) being invertible, \ie the input to the block can be calculated from the output; and (iii) the Jacobian of each block must be tractable so that the log probability density can be computed. These constraints can lead to trade-offs in which model expressiveness is sacrificed for tractability. 
Diffusion models do not have such constraints on their structures and may therefore be more suitable to model data priors.

\noindent
\textbf{Denoising Diffusion Models}
DDMs~\cite{sohl2015deep,Ho2020DDPM,nichol2021improved} are powerful generative models that learn to estimate gradients of the log data distribution.
% A stochastic process is defined in which a sample from the data distribution is progressively corrupted by repeated addition of Gaussian noise, and a model is trained to reverse that process by predicting and undoing the noise.
Once trained, Langevin dynamics sampling~\cite{welling2011bayesian} can be used to generate novel samples by performing a sequence of denoising steps starting from a random sample of a standard Gaussian distribution.
Denoising Diffusion Models have successfully been used to learn and sample images~\cite{Ho2020DDPM,song2021scorebased}, video~\cite{ho2022video}, speech~\cite{chen2020wavegrad,kong2020diffwave}, \etc. Recently,
multiple DDM-based models were proposed for the task of text-to-image synthesis, \eg DALL-E 2~\cite{ramesh2022hierarchical} and Imagen~\cite{saharia2022photorealistic}. Concurrently to our work, Dreamfusion~\cite{poole2022dreamfusion} has incorporated Imagen into NeRF optimization to generate novel 3D assets from a text input. Unlike our work, they use DDMs to guide optimization of NeRFs to match input text, while we use DDMs to regularize NeRFs given input training images.

\section{Method}\label{sec:method}

We start by covering preliminaries like NeRF and DDM training. Next, we describe the relation of DDMs to the gradient of the log-likelihood of the data, and show how we incorporate DDMs as NeRF regularizers. An overview of the our method is shown in Fig.~\ref{fig:pipeline}.

%------------------
\subsection{NeRFs}

Given a set of images of a scene $\mathcal{I}$ with camera intrinsic parameters and poses, we are interested in optimizing a density field $\sigma: \mathbb{R}^3 \rightarrow \mathbb{R}_{+}$ and color field $\mathbf{c}:\mathbb{R}^3 \times \mathbb{S}^2 \rightarrow \mathbb{R}_{[0, 1]}^3$, where the density field can be evaluated at any 3D coordinate $(x, y, z) \in \mathbb{R}^3$ and the color field can be evaluated at any 3D coordinate and viewing direction $\mathbf{d} \in \mathbb{S}^2$.

The density and color fields can be used to synthesize views of the scene from arbitrary cameras using differentiable rendering techniques. The expected color $C(\mathbf{r})$ of a ray $\mathbf{r}(t) = \mathbf{o} + t\mathbf{d}$ can be estimated using discrete samples $t_{0:N}$ (where $t_{i+1} > t_i > 0$), so 
\begin{equation}
\mathbf{C}(\mathbf{r}) \approxeq \sum_{i=1}^N w_i \mathbf{c}(\mathbf{r}(t_i), \mathbf{d}) + \left( 1 - \sum_{i=1}^N w_i \right) \mathbf{c}_\text{bg}
,
\end{equation}
where the weights of color contributions are
${w_i = T(t_i) \rho(t_i)}$, defined with
\begin{equation}
    \rho(t_i) = 1 - \exp( -\sigma(\mathbf{r}(t_i))(t_{i+1} - t_{i}))
\end{equation}
and
\begin{equation}
    T(t_i) = \prod_{j=1}^{i-1} (1 - \rho(t_j))
\end{equation}
is the accumulated transmittance function, \ie the probability of the ray $\mathbf{r}(t)$ starting at camera center $\mathbf{o}$ and reaching coordinate $\mathbf{r}(t_i)$ without being absorbed. The $\mathbf{c}_\text{bg}$ is the background color, which we set to white.

Similarly, one can compute the expected depth as 
\begin{equation}
\mathbf{D}(\mathbf{r}) = \frac{\sum_{i=1}^N w_i t_i}{\sum_{i=1}^N w_i}
.
\end{equation}
% \begin{equation}
% \mathbf{D}(\mathbf{r}) = \frac{\sum_{i=1}^N w_i t_i}{\sum_{i=1}^N w_i}
% .
% \end{equation}

The density and color fields are optimized to reduce the photometric reconstruction loss, \eg the L2 difference between input images and renderings from the same views is
\begin{equation}
    \mathcal{L}_{\text{photo}}(\sigma, \mathbf{c}) = \sum_{i=1}^\mathcal{I} ||I_i - \mathbf{C}_i||_2.
\label{eq:photo}
\end{equation}

The weights of color contributions $w_i$ in Eq.~\ref{eq:photo} can be regularized to have compact distribution~\cite{barron2022mipnerf360}:
\begin{align}
    \mathcal{L}_{\text{dist}} = \frac{1}{D(\mathbf{r})} &\biggr(\sum_{i, j} w_i w_k \left| \frac{t_i + t_{i+1}}{2} - \frac{t_j + t_{j+1}}{2} \right| \nonumber \\
    &+ \frac{1}{3} \sum_{i=1}^N w_i^2 (t_{i+1} - t_i)\biggr) ,
\end{align}
where we deviate from the original formulation by dividing through by the expected depth for the ray, which has the effect of increasing the strength of this regularizer for geometry that is close to the camera.

We also encourage the weights to sum to unity, because in real scenes we always expect a ray to be absorbed fully by the scene geometry:
\begin{equation}
    \mathcal{L}_{\text{fg}} = \left(1 - \sum_{i=1}^N w_i \right) ^ 2.
\end{equation}

In the few-view case, NeRFs frequently collapse to a degenerate solution in which each camera is fully or partially ``covered up'' with a copy of the corresponding training image. To prevent this, we introduce a regularization approach in which the placement of density that is contained in only one view frustum is penalized as
\begin{equation}
    \mathcal{L}_\text{fr} = \sum_i w_i \mathbf{1}(n_i <= 1),
\end{equation}
where $n_i$ is the number of training view frustums in which the point along the ray $\mathbf{r}(t_i)$ is contained, so that only weights which lie in fewer than two training frustums are included in the sum. This reflects our prior that most of the scene should be within the frustum of more than one of the training views.

Combining these geometric regularizers into a loss function already gives a very strong baseline,
\begin{equation}
    \mathcal{L}_\text{geom} = 
    \mathcal{L}_{\text{photo}}
    + \lambda_{\text{fg}} \mathcal{L}_{\text{fg}} 
    + \lambda_{\text{fr}} \mathcal{L}_{\text{fr}} 
    + \lambda_{\text{dist}} \mathcal{L}_{\text{dist}}.
    \label{eq:geomloss}
\end{equation}
The $\lambda$ coefficients control the contributions of the regularizers. In our experiments we refer to this combination of losses as our ``geometric baseline''.

\subsection{Score functions and DDMs}

Per Bayes' theorem, the \emph{a posteriori} probability of density and color fields given training views $\mathcal{I}$ is
\begin{equation}
    p(\sigma, \mathbf{c} | \mathcal{I}) \propto p(\mathcal{I} | \sigma, \mathbf{c}) p(\sigma, \mathbf{c}),
\end{equation}
where we drop the normalizing constant since it depends only on $\mathcal{I}$.
The log-posterior is
\begin{equation}
    \log(p(\mathcal{I} | \sigma, \mathbf{c})) + \log(p(\sigma, \mathbf{c})).
    \label{eq:posterior}
\end{equation}

% A prior over $(\sigma, \mathbf{c})$ can be modelled in the final loss formulation, \eg depth smoothness regularization, or the weak inductive bias associated with the chosen representation of the density and color fields, \eg MLP.

In practice, we are interested in maximizing $p(\sigma, \mathbf{c} | \mathcal{I})$ with stochastic gradient descent, which only requires computation of the gradient of the log-likelihood 
$\nabla_{\sigma, \mathbf{c}} \log(p(\mathcal{I} | \sigma, \mathbf{c}))$ and the gradient of the log-prior $\nabla_{\sigma, \mathbf{c}} \log(p(\sigma, \mathbf{c}))$, \ie the score function. Notice that explicit computation of the probabilities of the density and color fields $p(\sigma, \mathbf{c})$ is not required. Below, we describe how DDMs are learned and their relation to the score function.

\begin{figure}[ht]
  \centering
  \includegraphics[width=0.92\linewidth]{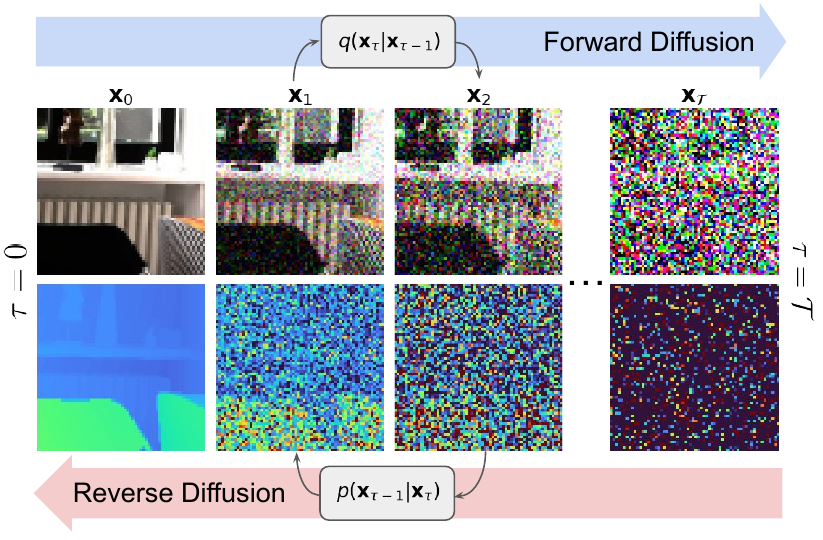}
  \\
  \centering
  (a)
  \\
  %%%%%%
  %%%%%%
  %%%%%%
  %%%%%%
\begin{minipage}[c]{0.9\linewidth}
\centering
\begin{minipage}[c]{\linewidth}
\centering
  \begin{minipage}[c]{0.09\linewidth}
  \includegraphics[width=\linewidth]{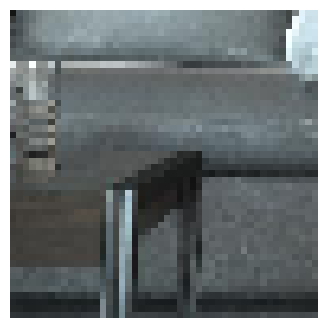}
  \end{minipage}
  \begin{minipage}[c]{0.09\linewidth}
  \includegraphics[width=\linewidth]{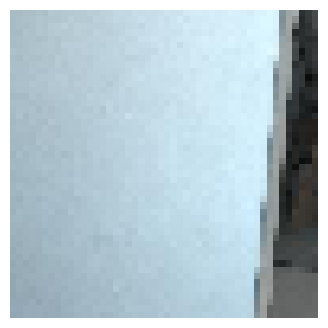}
  \end{minipage}
  \begin{minipage}[c]{0.09\linewidth}
  \includegraphics[width=\linewidth]{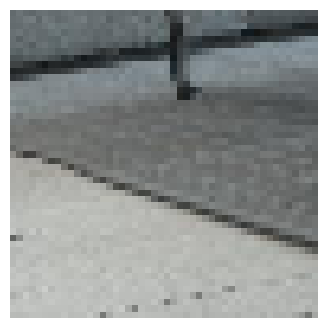}
  \end{minipage}
  \begin{minipage}[c]{0.09\linewidth}
  \includegraphics[width=\linewidth]{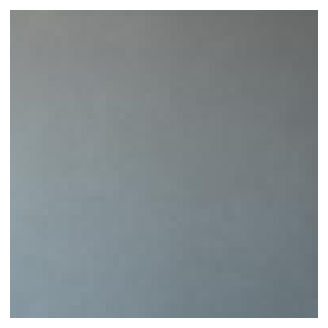}
  \end{minipage}
  \begin{minipage}[c]{0.09\linewidth}
  \includegraphics[width=\linewidth]{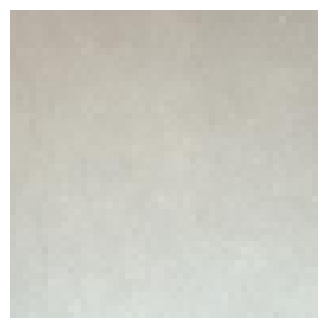}
  \end{minipage}
  \begin{minipage}[c]{0.09\linewidth}
  \includegraphics[width=\linewidth]{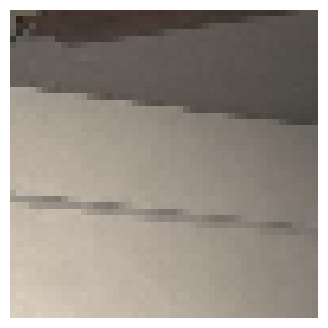}
  \end{minipage}
  \begin{minipage}[c]{0.09\linewidth}
  \includegraphics[width=\linewidth]{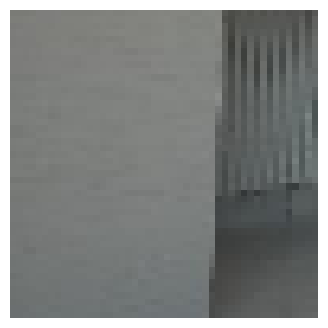}
  \end{minipage}
  \begin{minipage}[c]{0.09\linewidth}
  \includegraphics[width=\linewidth]{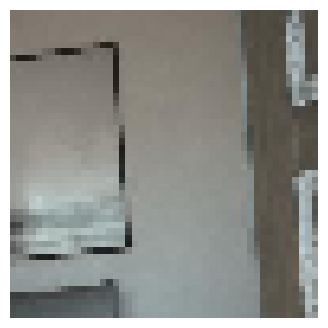}
  \end{minipage}
  \begin{minipage}[c]{0.09\linewidth}
  \includegraphics[width=\linewidth]{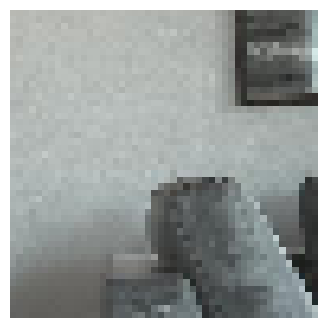}
  \end{minipage}
  \begin{minipage}[c]{0.09\linewidth}
  \includegraphics[width=\linewidth]{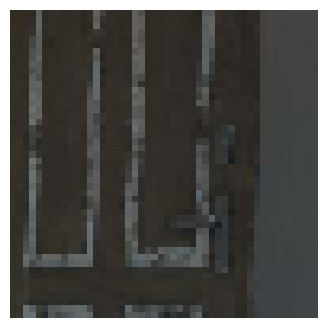}
  \end{minipage}
\end{minipage}
\\
\begin{minipage}[c]{\linewidth}
\centering
  \begin{minipage}[c]{0.09\linewidth}
  \includegraphics[width=\linewidth]{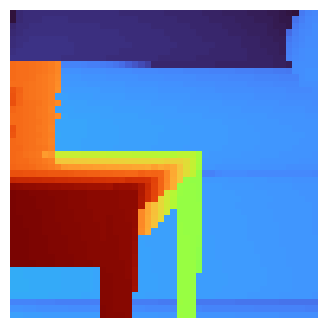}
  \end{minipage}
  \begin{minipage}[c]{0.09\linewidth}
  \includegraphics[width=\linewidth]{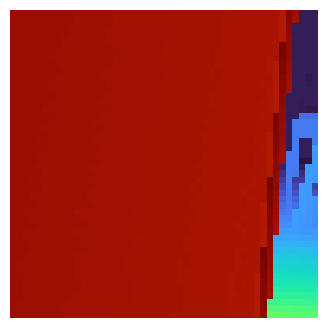}
  \end{minipage}
  \begin{minipage}[c]{0.09\linewidth}
  \includegraphics[width=\linewidth]{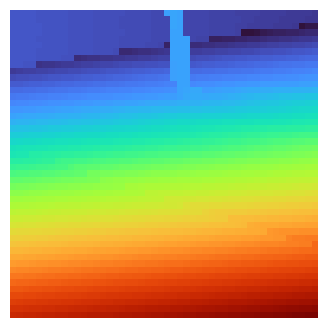}
  \end{minipage}
  \begin{minipage}[c]{0.09\linewidth}
  \includegraphics[width=\linewidth]{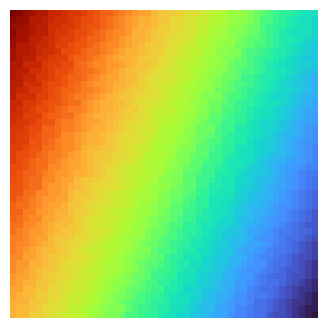}
  \end{minipage}
  \begin{minipage}[c]{0.09\linewidth}
  \includegraphics[width=\linewidth]{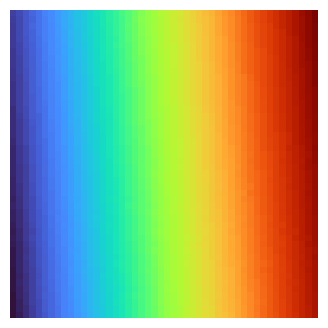}
  \end{minipage}
  \begin{minipage}[c]{0.09\linewidth}
  \includegraphics[width=\linewidth]{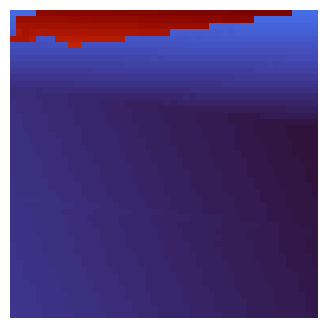}
  \end{minipage}
  \begin{minipage}[c]{0.09\linewidth}
  \includegraphics[width=\linewidth]{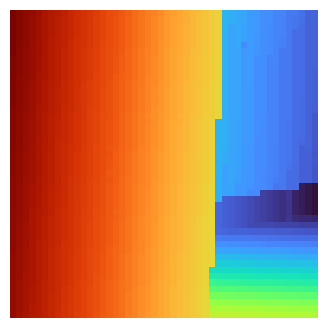}
  \end{minipage}
  \begin{minipage}[c]{0.09\linewidth}
  \includegraphics[width=\linewidth]{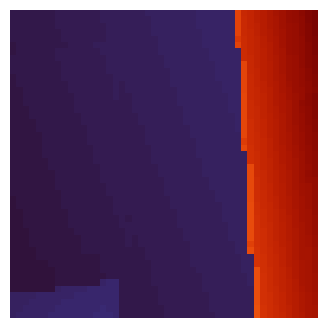}
  \end{minipage}
  \begin{minipage}[c]{0.09\linewidth}
  \includegraphics[width=\linewidth]{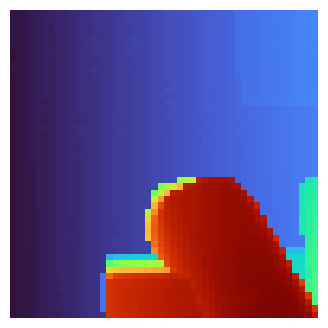}
  \end{minipage}
  \begin{minipage}[c]{0.09\linewidth}
  \includegraphics[width=\linewidth]{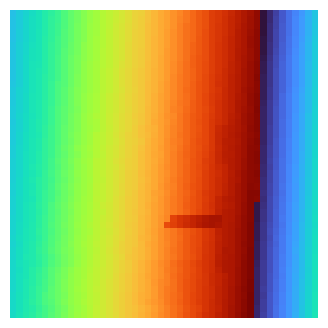}
  \end{minipage}
\end{minipage}
\end{minipage}
  \\
  \centering
  (b)
  \\
\vspace{2pt}
%%%%%%%%%%%%%%%%
%%%%%%%%%%%%%%%%
%%%%%%%%%%%%%%%%
%%%%%%%%%%%%%%%%
%%%%%%%%%%%%%%%%
%%%%%%%%%%%%%%%%
\begin{minipage}[c]{0.9\linewidth}
\centering
\begin{minipage}[c]{\linewidth}
\centering
  \begin{minipage}[c]{0.09\linewidth}
  \includegraphics[width=\linewidth]{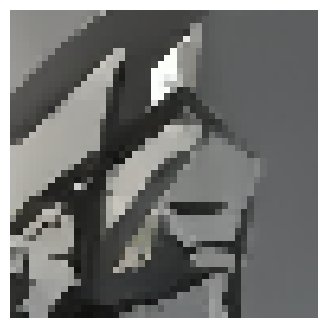}
  \end{minipage}
  \begin{minipage}[c]{0.09\linewidth}
  \includegraphics[width=\linewidth]{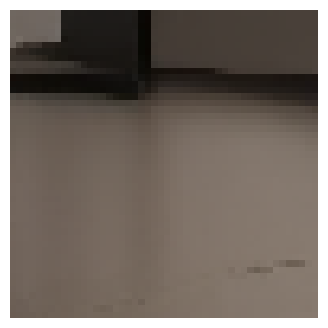}
  \end{minipage}
  \begin{minipage}[c]{0.09\linewidth}
  \includegraphics[width=\linewidth]{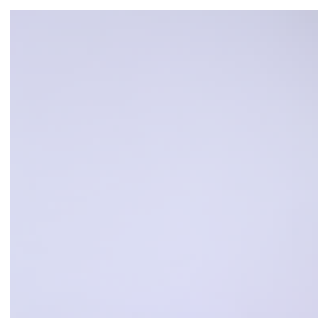}
  \end{minipage}
  \begin{minipage}[c]{0.09\linewidth}
  \includegraphics[width=\linewidth]{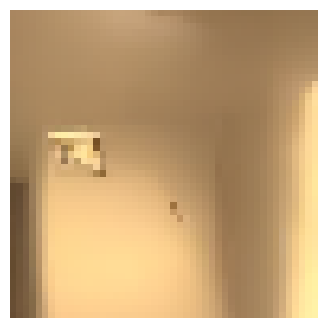}
  \end{minipage}
  \begin{minipage}[c]{0.09\linewidth}
  \includegraphics[width=\linewidth]{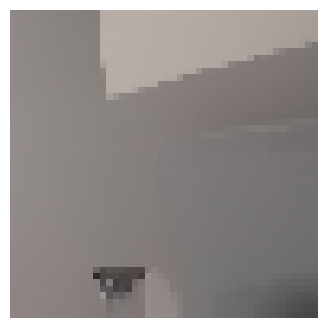}
  \end{minipage}
  \begin{minipage}[c]{0.09\linewidth}
  \includegraphics[width=\linewidth]{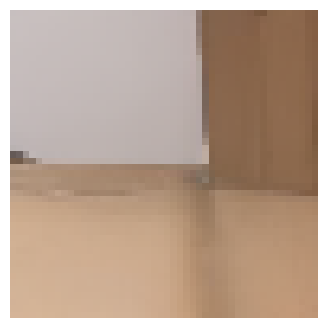}
  \end{minipage}
  \begin{minipage}[c]{0.09\linewidth}
  \includegraphics[width=\linewidth]{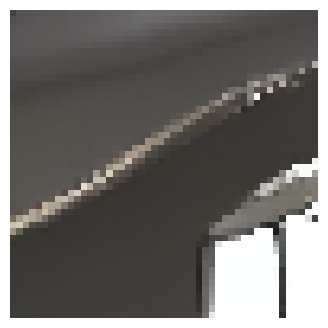}
  \end{minipage}
  \begin{minipage}[c]{0.09\linewidth}
  \includegraphics[width=\linewidth]{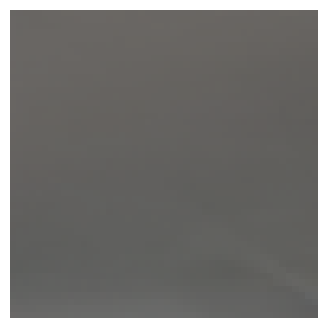}
  \end{minipage}
  \begin{minipage}[c]{0.09\linewidth}
  \includegraphics[width=\linewidth]{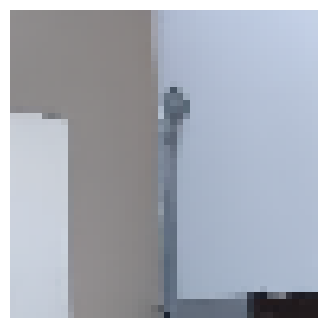}
  \end{minipage}
  \begin{minipage}[c]{0.09\linewidth}
  \includegraphics[width=\linewidth]{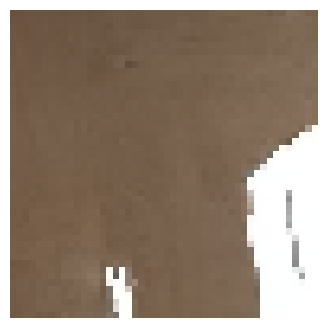}
  \end{minipage}
\end{minipage}
\\
\begin{minipage}[c]{\linewidth}
\centering
  \begin{minipage}[c]{0.09\linewidth}
  \includegraphics[width=\linewidth]{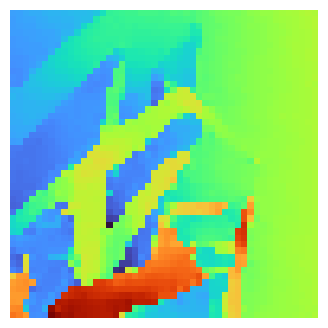}
  \end{minipage}
  \begin{minipage}[c]{0.09\linewidth}
  \includegraphics[width=\linewidth]{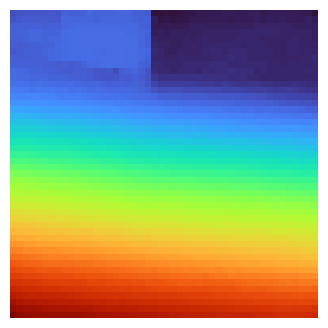}
  \end{minipage}
  \begin{minipage}[c]{0.09\linewidth}
  \includegraphics[width=\linewidth]{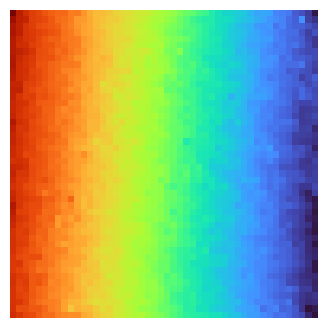}
  \end{minipage}
  \begin{minipage}[c]{0.09\linewidth}
  \includegraphics[width=\linewidth]{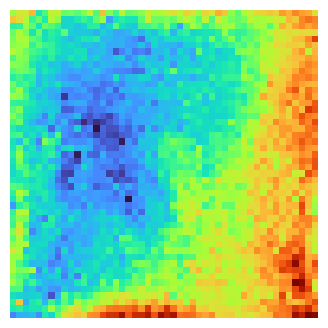}
  \end{minipage}
  \begin{minipage}[c]{0.09\linewidth}
  \includegraphics[width=\linewidth]{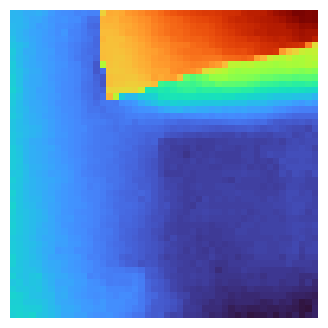}
  \end{minipage}
  \begin{minipage}[c]{0.09\linewidth}
  \includegraphics[width=\linewidth]{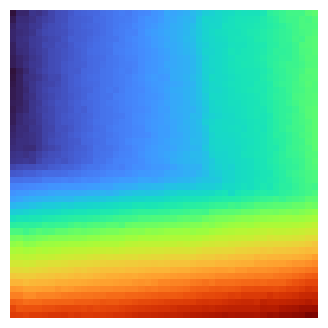}
  \end{minipage}
  \begin{minipage}[c]{0.09\linewidth}
  \includegraphics[width=\linewidth]{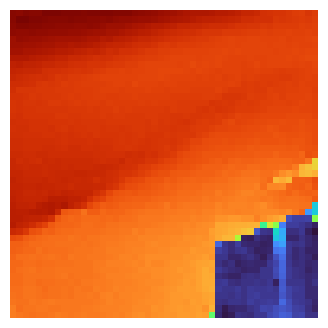}
  \end{minipage}
  \begin{minipage}[c]{0.09\linewidth}
  \includegraphics[width=\linewidth]{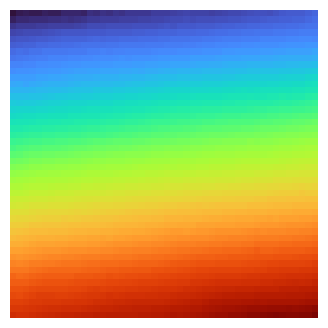}
  \end{minipage}
  \begin{minipage}[c]{0.09\linewidth}
  \includegraphics[width=\linewidth]{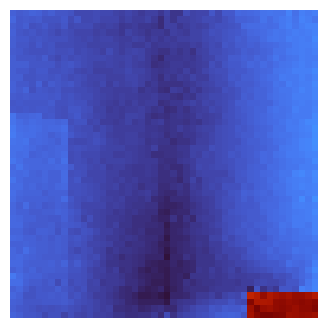}
  \end{minipage}
  \begin{minipage}[c]{0.09\linewidth}
  \includegraphics[width=\linewidth]{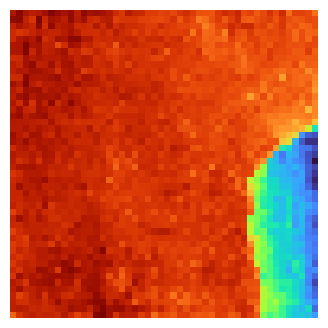}
  \end{minipage}
\end{minipage}
\end{minipage}
  \\
  \centering
  (c)
   \caption{(a) Illustration of forward and reverse diffusion processes.
   (b) Example RGBD patches in the training set of the DDM model extracted from Hypersim dataset.
   (c) Example RGBD patches generated with our DDM model trained on Hypersim dataset.
   Depths are shown as normalized inverse depths for visualization purposes. The noise in the samples is due to noise that is injected during the sampling process.
   }
   \label{fig:ddm_samples}
\end{figure}

The forward diffusion process progressively adds small Gaussian noise to a data sample $\mathbf{x}_0 \sim q(\mathbf{x})$ to produce progressively noisier versions, so
\begin{equation}
\mathbf{x}_\tau = \sqrt{\alpha_\tau}\mathbf{x}_{\tau-1} + \sqrt{\beta_\tau} \epsilon_{\tau-1},
\end{equation}
where $\epsilon_{\tau-1} \sim \mathcal{N}(\mathbf{0}, \mathbf{I})$ and 
$\alpha_\tau = 1 - \beta_\tau$, \ie the variances $\{\beta_{\tau} \}^{\mathcal{T}}_{\tau=1}$ control the noise schedule.
As the noise function is Gaussian, it follows from the reparameterization trick that
\begin{equation}
    q(\mathbf{x}_\tau | \mathbf{x}_{0}) = \mathcal{N}(\mathbf{x}_\tau; \sqrt{\bar{\alpha}_\tau} \mathbf{x}_{0}, (1-\bar{\alpha}_\tau \mathbf{I})),
\end{equation}
where $\bar{\alpha}_\tau = \prod_{s=0}^\tau \alpha_s$, allowing efficient generation of noised samples for arbitrary $\tau$. As $\mathcal{T} \rightarrow \infty$ the distribution of noised samples $\mathbf{x}_{\mathcal{T}}$ is equivalent to an isotropic unit Gaussian.

The DDM~\cite{sohl2015deep,Ho2020DDPM,nichol2021improved} is tasked to learn the reverse diffusion process:
\begin{equation}
    p(\mathbf{x}_{\tau-1} | \mathbf{x}_\tau) = \mathcal{N}(\mathbf{x}_{\tau-1}; \mathbf{\mu}(\mathbf{x}_\tau, \tau),
    \tilde{\beta}_\tau \mathbf{I})),
\end{equation}
where
$
    \tilde{\beta}_\tau = (1 - \bar{\alpha}_{\tau-1})\beta_\tau/(1 - \bar{\alpha}_\tau)
$
.

Since $\mathbf{x}_\tau$ is available as input to $\mathbf{\mu}(\mathbf{x}_\tau, \tau)$, the mean $\mathbf{\mu}(\mathbf{x}_\tau, \tau)$ can be computed by predicting noise $\epsilon_{\tau-1}$ from the noised input~\cite{Ho2020DDPM}:
\begin{equation}
    \mathbf{\mu}(\mathbf{x}_\tau, \tau) = \frac{1}{\sqrt{\alpha_\tau}}
    \left(
    \mathbf{x}_\tau -
    \frac{\beta_\tau}{\sqrt{1 - \bar{\alpha}_\tau}}
    \epsilon_\theta(\mathbf{x}_\tau, \tau)
    \right),
\end{equation}
using a neural network $\epsilon_\theta(\mathbf{x}_\tau, \tau)$.
 
Thus, one can learn the reverse diffusion process by training a neural network $\epsilon_\theta(\mathbf{x}_\tau, \tau)$ to estimate noise given a noised input and noise-level using the loss function:
\begin{equation}
    \mathbb{E}_{\mathbf{x}_0, \epsilon}
    \left[
    \frac{\beta_\tau}{2\alpha_\tau(1-\bar{\alpha}_\tau)}
    ||
    \epsilon - \epsilon_\theta \left(
    \sqrt{\bar{\alpha}_\tau} \mathbf{x}_0
    +
    \sqrt{1 - \bar{\alpha}_\tau} \epsilon,
    \tau
    \right)
    ||
    \right]
    ,
\end{equation}
where $\epsilon \sim \mathcal{N}(\mathbf{0}, \mathbf{I})$.
Fig.~\ref{fig:ddm_samples} (a) illustrates the forward and backwards processes.

Importantly, it was shown in~\cite{Vincent2011ACB,Ho2020DDPM} that a DDM noise estimator has a connection to score matching~\cite{hyvarinen05a,song2020sliced,SongNCSN} and is proportional to the score function:
\begin{equation}
    \epsilon_\theta(\mathbf{x}_\tau, \tau) \propto -\nabla_\mathbf{x} \log p(\mathbf{x}).
    \label{eq:score}
\end{equation}
Hence, taking steps in the negative direction to the noise predicted by the model is equivalent to moving towards the modes of the data distribution.
This can be used to generate samples from the data distribution using Langevin dynamics~\cite{welling2011bayesian,Ho2020DDPM,SongNCSN}.

In this work, we want to use a DDM model as a score function estimator to regularize NeRF reconstructions according to eq.~\ref{eq:posterior}. Hence, we model a prior over $(\sigma, \mathbf{c})$ by modeling the score function over the distribution of RGBD patches $
\epsilon_\theta (\{ \mathbf{C}(\mathbf{r}), \mathbf{D}(\mathbf{r}) | \mathbf{r} \in P \} ),
$
where $P$ is a set of rays that pass through a random $48 \times 48$ patch of pixels cast from a random camera. To allow control of the magnitude of the gradients, we further normalize the output of $\epsilon_\theta (\{ \mathbf{C}(\mathbf{r}), \mathbf{D}(\mathbf{r}) | \mathbf{r} \in P \} )$, and refer to this regularization function as $\epsilon_\theta$ (see supplementary for details).

To train our DDM we use \emph{Hypersim}~\cite{robertsHyperesim}, a photorealistic synthetic dataset for indoor scene understanding with ground truth images and depth maps. Specifically, we sample $48\times 48$ patches of images and depth maps to generate training data for the DDM
(removing problematic images and scenes as per dataset instructions); see Fig.~\ref{fig:ddm_samples}(b) for examples. Fig.~\ref{fig:ddm_samples}(c) shows samples of RGBD patches generated by our DDM model. The quality of samples indicates that DDM successfully learns the data distribution of the RGBD Hypersim patches.

%------------------
\subsection{Regularizing NeRFs with DDMs}

The gradient of the log-posterior~(\ref{eq:posterior}), which forms our loss function, is
\begin{equation}
    \nabla \log p(\sigma, \mathbf{c} | \mathcal{I}) = \nabla \log p(\sigma, \mathbf{c}) + \nabla \log p(\mathcal{I} | \sigma, \mathbf{c}).
\end{equation}

By plugging (\ref{eq:score}) into the above, we can use a diffusion model as a prior over $(\sigma, \mathbf{c})$. For the second term on the RHS we use loss in eq~\ref{eq:geomloss}, resulting in the following gradient for our loss function:
\begin{equation}
    \nabla \mathcal{L} = 
    \nabla \mathcal{L}_{\text{photo}}
    + \lambda_{\text{fg}} \nabla \mathcal{L}_{\text{fg}} 
    + \lambda_{\text{fr}} \nabla \mathcal{L}_{\text{fr}} 
    + \lambda_{\text{dist}} \nabla \mathcal{L}_{\text{dist}} 
    - \lambda_{\text{DDM}} \epsilon_\theta ,
    \label{eq:lossgrad}
\end{equation}
where $\lambda_\text{DDM}$ controls the weight of the our regularizer.

During NeRF optimization we compute the gradient of the loss as per eq.~\ref{eq:lossgrad} and backpropagate as usual to obtain gradients for the NeRF density and color field parameters.

%------------------
\subsection{Implementation Details}
We use the training protocol of ~\cite{torch-ddm,Ho2020DDPM} to train our DDM model. We optimize the DDM for 650,000 steps with batch size 32 on 1 GPU.

\begin{table*}[ht]
  \centering
  \footnotesize
  \begin{tabular}{|p{0.7em}|l|c|*{4}{p{1.9em}p{1.9em}p{1.9em}|}}
  \cline{2-15}
  \multicolumn{1}{c|}{} &
  \multirow{2}{*}{Method} & 
  \multirow{2}{*}{\centering Setting} &
  \multicolumn{3}{c|}{PSNR $\uparrow$} &
  \multicolumn{3}{c|}{SSIM $\uparrow$} &
  \multicolumn{3}{c|}{LPIPS-VGG $\downarrow$} &
  \multicolumn{3}{c|}{Average $\downarrow$}
  \\
  \multicolumn{1}{c|}{} &
  &
  & {\tiny  3-view} 
  & {\tiny  6-view} 
  & {\tiny  9-view}
  & {\tiny  3-view} 
  & {\tiny  6-view} 
  & {\tiny  9-view}
  & {\tiny  3-view} 
  & {\tiny  6-view} 
  & {\tiny  9-view}
  & {\tiny  3-view} 
  & {\tiny  6-view} 
  & {\tiny  9-view}
  \\
\hline
% \arrayrulecolor{black}
% *NGP~\cite{mueller2022instant}
% & Optimized per Scene 
% & 16.15 & 20.22 & 21.86 
% & 0.376 & 0.590 & 0.685 
% & 0.450 & 0.241 & 0.147 
% & 0.210 & 0.128 & 0.090 
% \\

\multirow{7}{*}{
\rotatebox[origin=c]{90}{
  \parbox[c]{0.7cm}{\centering
  LLFF
  }}
}
&
\cellcolor{LightBlue}
mip-NeRF~\cite{barron2021mip}
& {\scriptsize Optimized per Scene}
& 14.62 & 20.87 & 24.26
& 0.351 & 0.692 & \underline{0.805}
& 0.495 & 0.255 & \underline{0.172}
& 0.246 & 0.114 & 0.073
\\

&
\cellcolor{LightBlue}
DietNeRF~\cite{Jain_2021_ICCV}
& \cellcolor{LightGray}{\scriptsize Optimized per Scene}
& \cellcolor{LightGray}  14.94 & \cellcolor{LightGray}  21.75 & \cellcolor{LightGray}  24.28
& \cellcolor{LightGray}  0.370 & \cellcolor{LightGray}  0.717 & \cellcolor{LightGray}  0.801
& \cellcolor{LightGray}  0.496 & \cellcolor{LightGray}  0.248 & \cellcolor{LightGray}  0.183
& \cellcolor{LightGray}  0.240 & \cellcolor{LightGray}  0.105 & \cellcolor{LightGray}  0.073
\\

&
\cellcolor{LightBlue}
PixelNeRF ft ~\cite{yu2021pixelnerf}
& {\scriptsize DTU + ft per Scene}
& 16.17 & 17.03 & 18.92
& 0.438 & 0.473 & 0.535
& 0.512 & 0.477 & 0.430
& 0.217 & 0.196 & 0.163
\\

&
\cellcolor{LightBlue}
MVSNeRF ft ~\cite{mvsnerf}
& \cellcolor{LightGray}{\scriptsize DTU + ft per Scene}
& \cellcolor{LightGray}  17.88 & \cellcolor{LightGray}  19.99 & \cellcolor{LightGray}  20.47
& \cellcolor{LightGray}  0.584 & \cellcolor{LightGray}  0.660 & \cellcolor{LightGray}  0.695
& \cellcolor{LightGray}  \underline{0.327} & \cellcolor{LightGray}  0.264 & \cellcolor{LightGray}  0.244
& \cellcolor{LightGray}  0.157 & \cellcolor{LightGray}  0.122 & \cellcolor{LightGray}  0.111
\\

&
\cellcolor{LightBlue}
RegNeRF~\cite{Niemeyer2021Regnerf}
& {\scriptsize Optimized per Scene}
& 19.08 & 21.10 & 24.86
& \underline{0.587} & \underline{0.760} & \textbf{0.820}
& 0.336 & \textbf{0.206} & \textbf{0.161}
& 0.146 & 0.086 & \underline{0.067}
\\

&
\cellcolor{LightBlue}
Geometric Baseline
& \cellcolor{LightGray}{\scriptsize Optimized per Scene}
& \cellcolor{LightGray}  \textbf{19.88} & \cellcolor{LightGray}  \textbf{24.28} & \cellcolor{LightGray}  \textbf{25.10}
& \cellcolor{LightGray}  \textbf{0.590} & \cellcolor{LightGray}  \textbf{0.765} & \cellcolor{LightGray}  0.802
& \cellcolor{LightGray}  \textbf{0.312} & \cellcolor{LightGray}  \underline{0.210} & \cellcolor{LightGray}  0.189
& \cellcolor{LightGray}  \textbf{0.129} & \cellcolor{LightGray}  \textbf{0.076} & \cellcolor{LightGray}  \textbf{0.066}
\\

&
\cellcolor{LightBlue}
\textbf{DiffusioNeRF} (Ours)
& {\scriptsize Optimized per Scene}
& \underline{19.79} & \underline{23.79} & \underline{25.02} 
& 0.568 & 0.747 & 0.785 
& 0.338 & 0.237 & 0.212 
& \underline{0.136} & \underline{0.083} & 0.071
\\

% --------------------------------------------------
% ABOVE IS LLFF
% BELOW IS DTU
% --------------------------------------------------
\hline

\multirow{7}{*}{
\rotatebox[origin=c]{90}{
  \parbox[c]{0.7cm}{\centering
  DTU
  }}
} 
&
\cellcolor{LightRed}
mip-NeRF~\cite{barron2021mip}
& \cellcolor{LightGray}{\scriptsize Optimized per Scene}
& \cellcolor{LightGray}  8.68   & \cellcolor{LightGray}  16.54 & \cellcolor{LightGray}  23.58
& \cellcolor{LightGray}  0.571  & \cellcolor{LightGray}  0.741 & \cellcolor{LightGray}  0.879
& \cellcolor{LightGray}  0.353  & \cellcolor{LightGray}  0.198 & \cellcolor{LightGray}  \underline{0.092}
& \cellcolor{LightGray}  0.323  & \cellcolor{LightGray}  0.148 & \cellcolor{LightGray}  0.056
\\

&
\cellcolor{LightRed}
DietNeRF~\cite{Jain_2021_ICCV}
& {\scriptsize Optimized per Scene}
& 11.85 & 20.63 & 23.83
& 0.633 & 0.778 & 0.823
& 0.314 & 0.201 & 0.173
& 0.243 & 0.101 & 0.068
\\

&
\cellcolor{LightRed}
PixelNeRF ft ~\cite{yu2021pixelnerf}
& \cellcolor{LightGray}{\scriptsize DTU + ft per Scene}
& \cellcolor{LightGray}  \textbf{18.95} & \cellcolor{LightGray}  \underline{20.56} & \cellcolor{LightGray}  21.83
& \cellcolor{LightGray}  0.710 & \cellcolor{LightGray}  0.753 & \cellcolor{LightGray}  0.781
& \cellcolor{LightGray}  0.269 & \cellcolor{LightGray}  0.223 & \cellcolor{LightGray}  0.203
& \cellcolor{LightGray}  0.125 & \cellcolor{LightGray}  0.104 & \cellcolor{LightGray}  0.090
\\

&
\cellcolor{LightRed}
MVSNeRF ft ~\cite{mvsnerf}
& {\scriptsize DTU + ft per Scene}
& 18.54 & 20.49 & 22.22
& \textbf{0.769} & \underline{0.822} & 0.853
& \underline{0.197} & 0.155 & 0.135
& \underline{0.113} & 0.089 & 0.069
\\

&
\cellcolor{LightRed}
RegNeRF~\cite{Niemeyer2021Regnerf}
& \cellcolor{LightGray}{\scriptsize Optimized per Scene}
& \cellcolor{LightGray}  \underline{18.89} & \cellcolor{LightGray}  \textbf{22.20} & \cellcolor{LightGray}  \underline{24.93}
& \cellcolor{LightGray}  \underline{0.745} & \cellcolor{LightGray}  \textbf{0.841} & \cellcolor{LightGray}  \textbf{0.884}
& \cellcolor{LightGray}  \textbf{0.190} & \cellcolor{LightGray}  \textbf{0.117} & \cellcolor{LightGray}  \textbf{0.089}
& \cellcolor{LightGray}  \textbf{0.112} & \cellcolor{LightGray}  \textbf{0.071} & \cellcolor{LightGray}  \textbf{0.047}
\\

&
\cellcolor{LightRed}
Geometric Baseline
& {\scriptsize Optimized per Scene}
& 13.60 & 16.43 & 22.01
& 0.661 & 0.759 & 0.853
& 0.255 & 0.182 & 0.121
& 0.193 & 0.123 & 0.067
\\

&
\cellcolor{LightRed}
\textbf{DiffusioNeRF} (Ours)
& \cellcolor{LightGray}{\scriptsize Optimized per Scene}
& \cellcolor{LightGray}  16.20 & \cellcolor{LightGray}  20.34 & \cellcolor{LightGray}  \textbf{25.18}
& \cellcolor{LightGray}  0.698 & \cellcolor{LightGray}  0.818 & \cellcolor{LightGray}  \underline{0.883}
& \cellcolor{LightGray}  0.207 & \cellcolor{LightGray}  \underline{0.139} & \cellcolor{LightGray}  0.095
& \cellcolor{LightGray}  0.146 & \cellcolor{LightGray}  \underline{0.081} & \cellcolor{LightGray}  \textbf{0.047}
\\
\arrayrulecolor{gray}\hline

  \end{tabular}
  \caption{\textbf{DiffusioNeRF \vs SOTA} in novel view synthesis task on \colorbox{LightBlue}{LLFF} and \colorbox{LightRed}{DTU} datasets with few input views~\cite{yu2021pixelnerf,Niemeyer2021Regnerf}. We report scores on PSNR, SSIM, LPIPS and Average metrics averaged over all 8 scenes when NeRFs are fitted with 3, 6 and 9 training views. For each view/metric combination the \textbf{first} and \underline{second} scores are highlighted.
  }
  \label{tab:novel_view}
\end{table*}

We use the torch-ngp~\cite{torch-ngp} implementation of Instant NGP~\cite{mueller2022instant} with the tiny-cuda-nn~\cite{tiny-cuda-nn} back-end as the NeRF model for our experiments.
NeRFs are optimized for 12,000 steps, where the first 2500 steps are optimized with $\lambda_\text{dist}=0$ and the diffusion time parameter $\tau$ smoothly interpolates from 0.1 to 0, hence we set $\bar{\alpha}_\tau= \cos({0.5\pi(\tau+0.008)} / {1.008})$ and other variables are derived accordingly. By scheduling $\tau$ this way the diffusion model is conditioned to expect progressively less noisy inputs as the NeRF trains and generates increasingly more accurate colors and depths.
After 3000 steps, $\lambda_\text{dist}$ linearly increases from 0 until it reaches its maximum value at 8000 steps, where the maximum value is $1\times10^{-4}$ for the DTU dataset and $1.5\times10^{-5}$ for the LLFF dataset.
We empirically found that this schedule of $\tau$ and regularization weights produces best results. On a single Nvidia A100 GPU our NeRF model trains in approximately 30 minutes per scene.

Furthermore, 25\% of the time we use a training pose for patch rendering, and sample the RGB component of the RGBD patch directly from the training image. This is helpful in the early stages, when NeRF renderings are not yet accurate.

%------------------------------------------------------------------------
\section{Experiments}\label{sec:experiments}

\noindent
\textbf{Datasets}
We experiment on two datasets: LLFF and DTU.

The \emph{LLFF}~\cite{mildenhall2019llff} dataset has 8 scenes with 20-62 images per scene captured with a handheld camera. The scenes are reconstructed with COLMAP~\cite{schoenberger2016sfm} to estimate camera intrinsics, camera poses and the 3D bounds of the scenes. 
A few images are used for training and test images are used to evaluate novel view synthesis quality. We select LLFF for evaluations as it allows comparison against other SOTA NeRF models, such as RegNeRF~\cite{Niemeyer2021Regnerf}.

The \emph{DTU}~\cite{jensen2014large} dataset consists of images of objects placed on a table against black background. Images and depth maps are captured with structured light scanner mounted on an industrial robot arm. The dataset provides images, poses, and ground truth point clouds for evaluation.

For novel-view synthesis with few view setting on DTU, we use the test set of 15 scans of PixelNeRF~\cite{yu2021pixelnerf}, allowing comparison against other methods.

We use the test set of 15 scans defined in ~\cite{Oechsle2021ICCV,yariv2021volume,Yu2022MonoSDF} to evaluate  geometry quality, \eg via the surface method of evaluation as described in UNISURF~\cite{Oechsle2021ICCV}. Traditionally, geometry estimated by the density field of a NeRF may not allow accurate surface reconstruction compared to occupancy and SDF-based approaches~\cite{Oechsle2021ICCV}, which score higher on DTU, \eg~\cite{Yu2022MonoSDF,Oechsle2021ICCV,yariv2020multiview,yariv2021volume}. % As our regularizer is designed to increase the quality of fitted geometry, we choose to measure improvements in the estimated surface geometry on DTU.

\noindent
\textbf{Metrics}
For the task of novel-view synthesis, hold-out views of the scene are used as ground truth to compare against synthesized views. Image similarity metrics such as 
PSNR, SSIM~\cite{wang2004image} and LPIPS~\cite{zhang2018unreasonable} are measured for each test view and average score per each scene is reported. We also report an ``Average'' score, specifically the geometric mean of the three metrics as per~\cite{barron2021mip}:
$\sqrt[3]{ 10^{-\text{PSNR}/10} \cdot \sqrt{1 - \text{SSIM}} \cdot \text{LPIPS} }$.

For the geometry estimation task, we convert an isosurface of the density field into a mesh using the marching cubes.
% We extract a mesh from the fitted NeRFs and use visibility culling to retain only those parts of the mesh that are actually visibility in at least one training view.
% We also use provided object masks to remove regions of the mesh corresponding to the background as opposed to the object of interest.
The mesh is culled to retain only parts that are visible in at least one training view and the background surfaces are masked out.
We then sample the mesh to generate a point cloud, and report the average chamfer $L1$ distance between the estimated and ground truth point clouds.

\begin{figure*}[ht]
\centering
\begin{minipage}[c]{0.95\textwidth}
\begin{minipage}[c]{\textwidth}
\centering
  \begin{minipage}[c]{0.13\linewidth}
\small
Ground Truth
\end{minipage}
  ~
  \begin{minipage}[c]{0.26\linewidth}
  % \centering 
  \small
  \hspace{35pt}
  RegNeRF
  \end{minipage}
  ~
  \begin{minipage}[c]{0.26\linewidth}
  \centering
  \small
  Geometric Baseline
  \end{minipage}
  ~
  \begin{minipage}[c]{0.26\linewidth}
  \hspace{25pt}
  \small
  DiffusioNeRF (Ours)
  \end{minipage}
\end{minipage}
\\
\begin{minipage}[c]{\textwidth}
\centering
  \begin{minipage}[c]{0.13\linewidth}
  \includegraphics[width=0.99\linewidth]{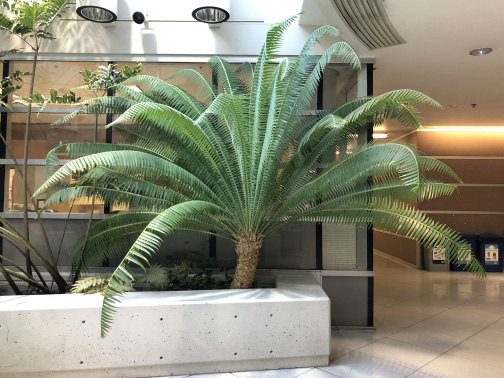}
  \end{minipage}
  ~
  \begin{minipage}[c]{0.13\linewidth}
  \includegraphics[width=0.99\linewidth]{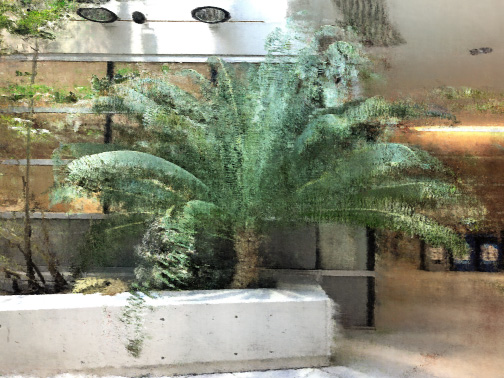}
  \end{minipage}
  ~
  \begin{minipage}[c]{0.13\linewidth}
  \includegraphics[width=0.99\linewidth]{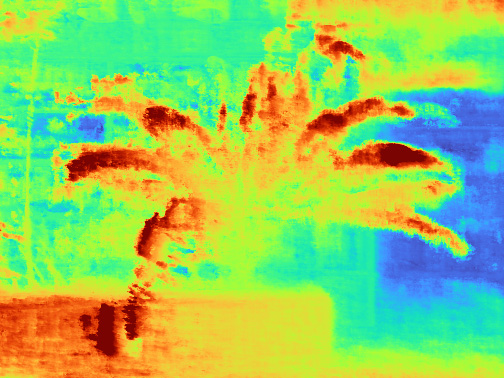}
  \end{minipage}
  ~
  \begin{minipage}[c]{0.13\linewidth}
  \includegraphics[width=0.99\linewidth]{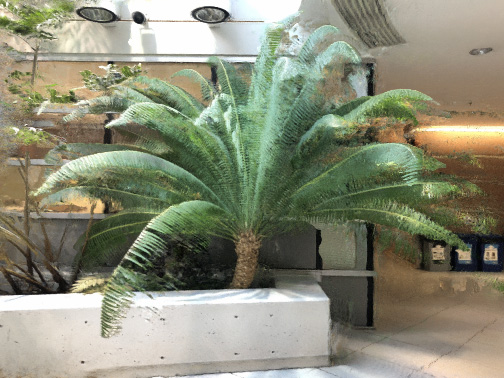}
  \end{minipage}
  ~
  \begin{minipage}[c]{0.13\linewidth}
  \includegraphics[width=0.99\linewidth]{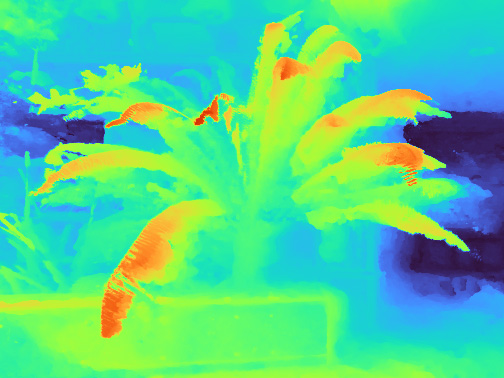}
  \end{minipage}
  ~
  \begin{minipage}[c]{0.13\linewidth}
  \includegraphics[width=0.99\linewidth]{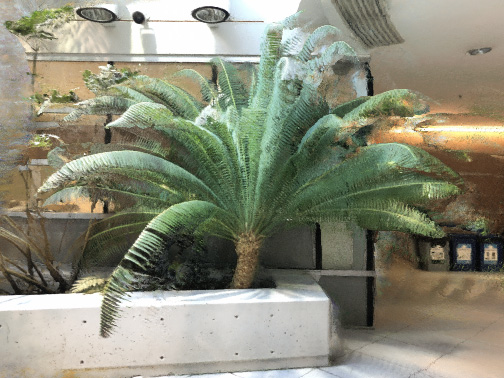}
  \end{minipage}
  ~
  \begin{minipage}[c]{0.13\linewidth}
  \includegraphics[width=0.99\linewidth]{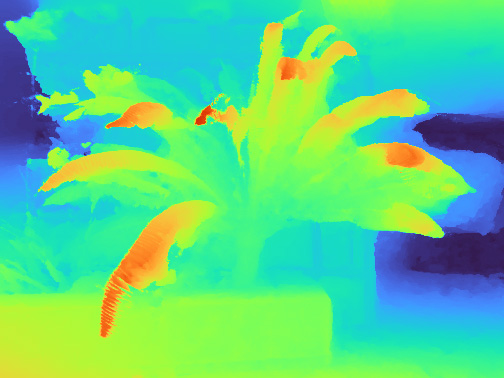}
  \end{minipage}
\end{minipage}
\\
\begin{minipage}[c]{\linewidth}
\centering
  \begin{minipage}[c]{0.13\linewidth}
  \includegraphics[width=0.99\linewidth]{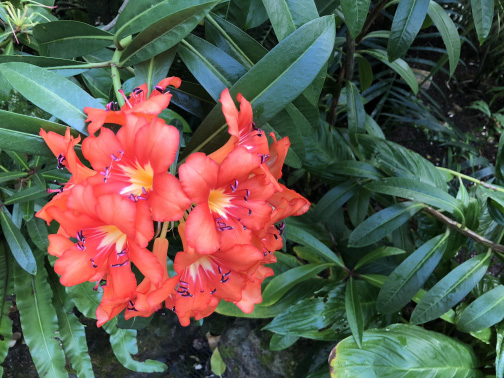}
  \end{minipage}
  ~
  \begin{minipage}[c]{0.13\linewidth}
  \includegraphics[width=0.99\linewidth]{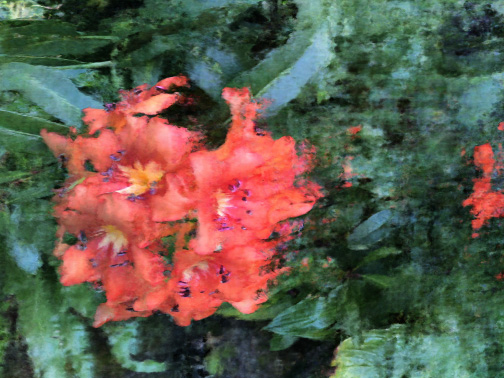}
  \end{minipage}
  ~
  \begin{minipage}[c]{0.13\linewidth}
  \includegraphics[width=0.99\linewidth]{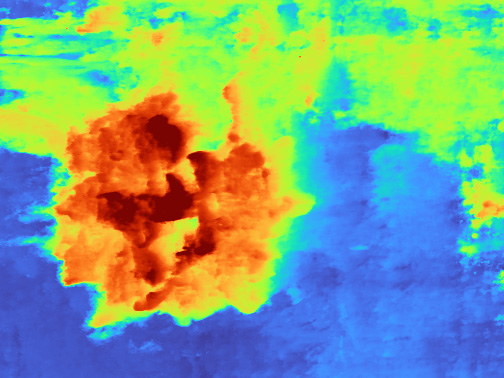}
  \end{minipage}
  ~
  \begin{minipage}[c]{0.13\linewidth}
  \includegraphics[width=0.99\linewidth]{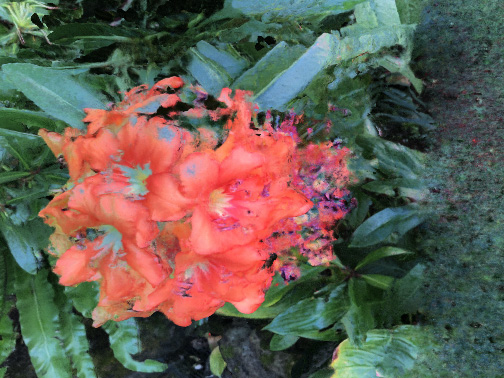}
  \end{minipage}
  ~
  \begin{minipage}[c]{0.13\linewidth}
  \includegraphics[width=0.99\linewidth]{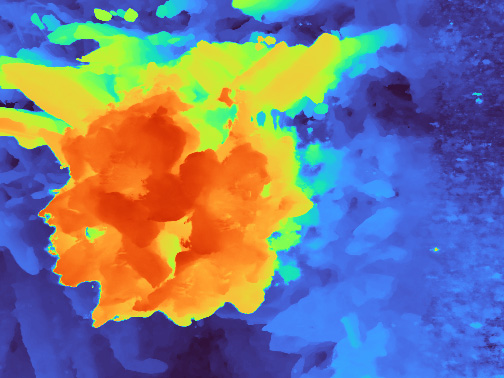}
  \end{minipage}
  ~
  \begin{minipage}[c]{0.13\linewidth}
  \includegraphics[width=0.99\linewidth]{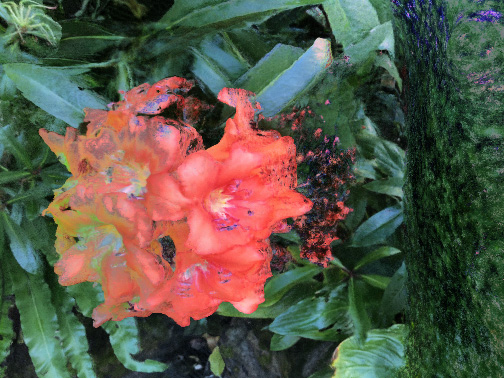}
  \end{minipage}
  ~
  \begin{minipage}[c]{0.13\linewidth}
  \includegraphics[width=0.99\linewidth]{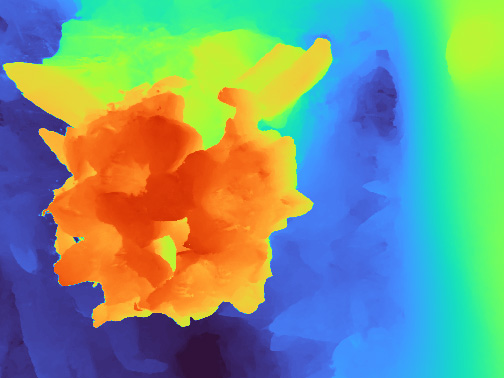}
  \end{minipage}
\end{minipage}
\\
\begin{minipage}[c]{\linewidth}
\centering
  \begin{minipage}[c]{0.13\linewidth}
  \includegraphics[width=0.99\linewidth]{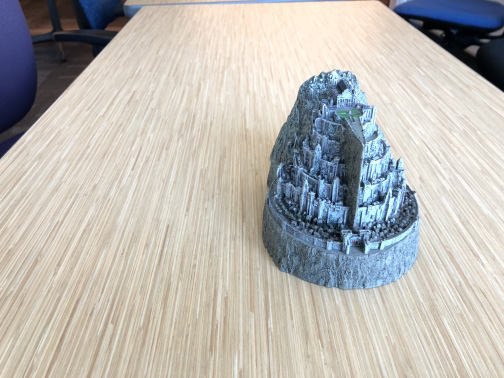}
  \end{minipage}
  ~
  \begin{minipage}[c]{0.13\linewidth}
  \includegraphics[width=0.99\linewidth]{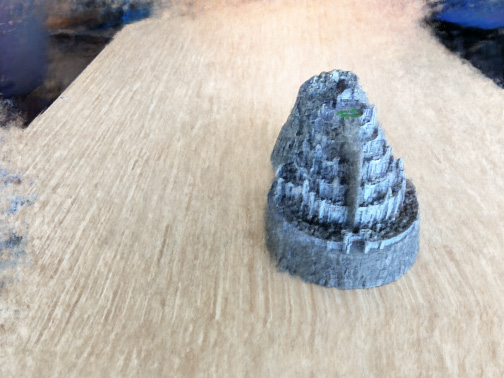}
  \end{minipage}
  ~
  \begin{minipage}[c]{0.13\linewidth}
  \includegraphics[width=0.99\linewidth]{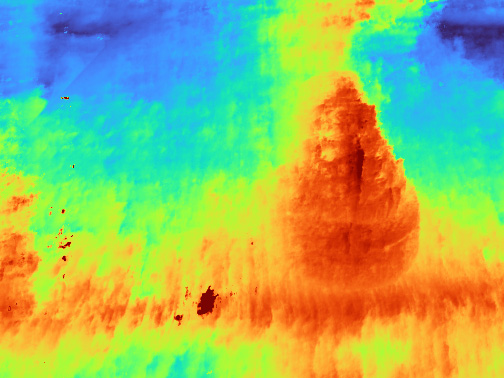}
  \end{minipage}
  ~
  \begin{minipage}[c]{0.13\linewidth}
  \includegraphics[width=0.99\linewidth]{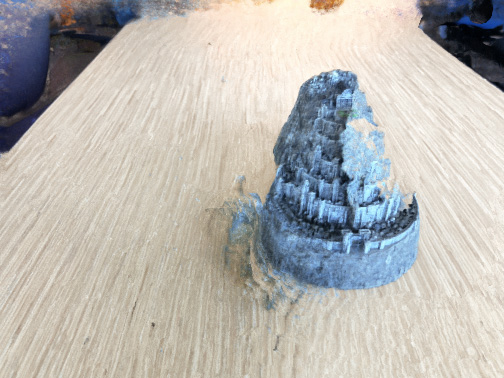}
  \end{minipage}
  ~
  \begin{minipage}[c]{0.13\linewidth}
  \includegraphics[width=0.99\linewidth]{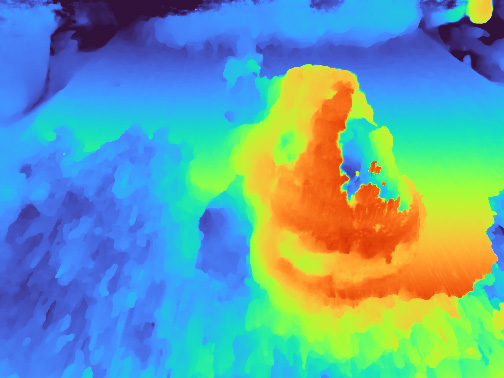}
  \end{minipage}
  ~
  \begin{minipage}[c]{0.13\linewidth}
  \includegraphics[width=0.99\linewidth]{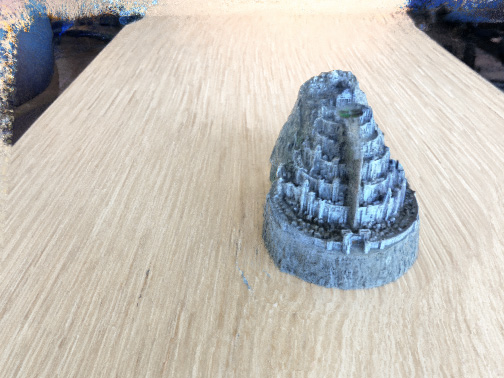}
  \end{minipage}
  ~
  \begin{minipage}[c]{0.13\linewidth}
  \includegraphics[width=0.99\linewidth]{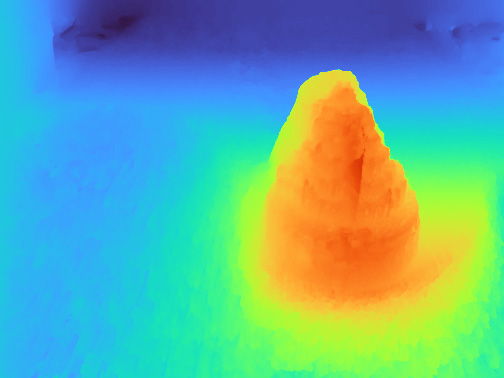}
  \end{minipage}
\end{minipage}
\\
\begin{minipage}[c]{\linewidth}
\centering
  \begin{minipage}[c]{0.13\linewidth}
  \includegraphics[width=0.99\linewidth]{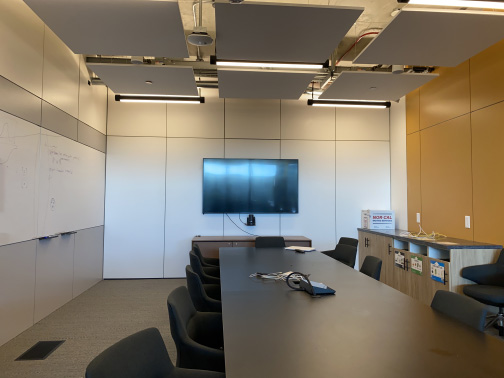}
  \end{minipage}
  ~
  \begin{minipage}[c]{0.13\linewidth}
  \includegraphics[width=0.99\linewidth]{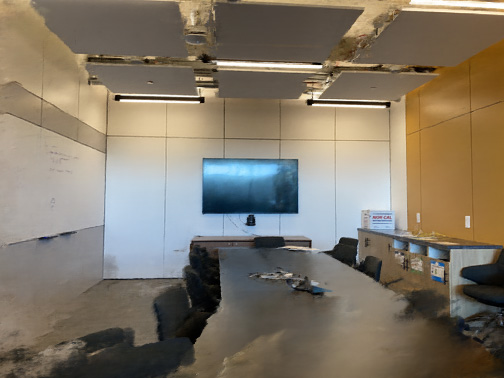}
  \end{minipage}
  ~
  \begin{minipage}[c]{0.13\linewidth}
  \includegraphics[width=0.99\linewidth]{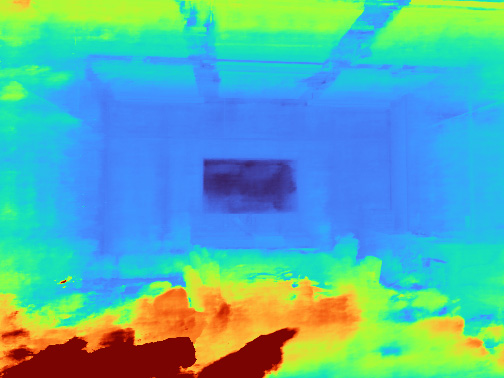}
  \end{minipage}
  ~
  \begin{minipage}[c]{0.13\linewidth}
  \includegraphics[width=0.99\linewidth]{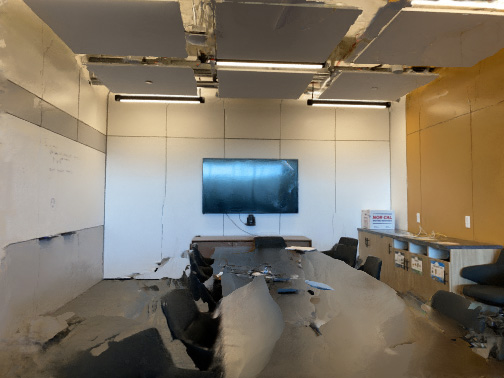}
  \end{minipage}
  ~
  \begin{minipage}[c]{0.13\linewidth}
  \includegraphics[width=0.99\linewidth]{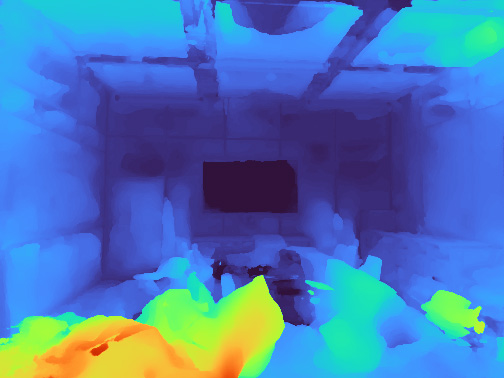}
  \end{minipage}
  ~
  \begin{minipage}[c]{0.13\linewidth}
  \includegraphics[width=0.99\linewidth]{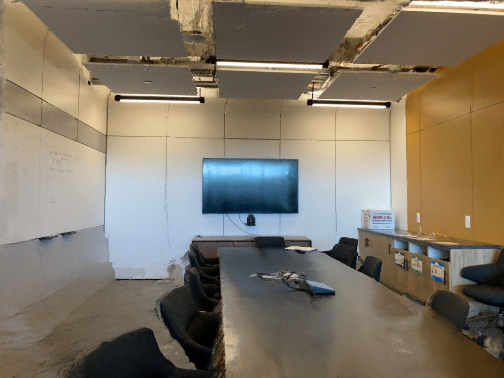}
  \end{minipage}
  ~
  \begin{minipage}[c]{0.13\linewidth}
  \includegraphics[width=0.99\linewidth]{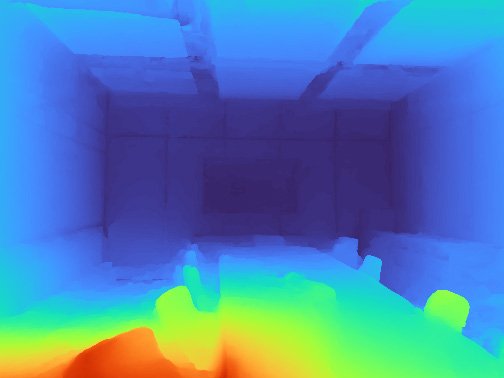}
  \end{minipage}
\end{minipage}
\end{minipage}
   \caption{Qualitative results for the task of novel view synthesis on LLFF dataset. NeRF models are trained with 3 views and rendered from one of test views.
   Our DDM model encourages more realistic geometry as seen in the depth maps.}
   \label{fig:qualitative_results_llff}
\end{figure*}

\subsection{Evaluations}
Table~\ref{tab:novel_view} show a comparison of our geometric baseline and our model against SOTA methods on LLFF and DTU datasets when trained with 3, 6 and 9 views. When the number of views is low, the regularizer can have a large impact on the final result, which allows easier comparison of regularizers.
As seen from table~\ref{tab:novel_view}, the geometric baseline and our method both score favorably to other methods, often achieving the best scores.
Our geometric baseline has higher metrics on LLFF, however there are artifacts in the generated test views that can be seen in Fig.~\ref{fig:qualitative_results_llff}. Our diffusion model-based method generates more plausible depths compared to the geometric baseline, see section~\ref{sec:ablation}. One side-effect is over-smoothing of thin-structures (\eg the top row in Fig.~\ref{fig:qualitative_results_llff}).
It is also noteworthy that test views contain parts of the scene that are not visible in any of the training views. These occluded parts of the scene can impact reconstruction scores significantly (see supplementary for details).

Table~\ref{tab:recon} shows an evaluation of reconstruction quality on 15 scans of the DTU dataset when NeRFs are fitted with all views. 
In the large number of views regime, the priors are less important as training views provide more information about the scene.  
Nevertheless, the priors should not introduce any undesirable artifacts and can help with ambiguous regions such as textureless table.
Despite DDM being trained on images of indoor room-sized scenes, it shows good generalization to the object-centric reconstruction task. Our density-based method performs adequately when compared to occupancy and SDF-based methods.

\begin{figure*}[ht]
\hspace{-0.3cm}
\begin{minipage}[c]{1.04\textwidth}
\begin{minipage}[c]{\linewidth}
\centering
  \begin{minipage}[c]{0.05\linewidth}
  \centering
  \small
  Scan \#
  \end{minipage}
  ~
  \begin{minipage}[c]{0.13\linewidth}
  \centering
  \small
  RGB
  \end{minipage}
  ~
  \begin{minipage}[c]{0.13\linewidth}
  \centering
  \small
  NeuS~\cite{wang2021neus}
  \end{minipage}
  ~
  \begin{minipage}[c]{0.13\linewidth}
  \centering
  \small
  VolSDF~\cite{yariv2021volume}
  \end{minipage}
  ~
  \begin{minipage}[c]{0.13\linewidth}
  \centering
  \small
  MonoSDF~\cite{Yu2022MonoSDF}
  \end{minipage}
  ~
  \begin{minipage}[c]{0.13\linewidth}
  \centering
  \small
  Geom. Baseline
  \end{minipage}
  ~
  \begin{minipage}[c]{0.13\linewidth}
  \centering
  \small
  Ours
  \end{minipage}
\end{minipage}
\\
\begin{minipage}[c]{\linewidth}
\centering
  \begin{minipage}[c]{0.05\linewidth}
  \centering
  \small
  24
  \end{minipage}
  ~
  \begin{minipage}[c]{0.13\linewidth}
  \includegraphics[width=0.99\linewidth]{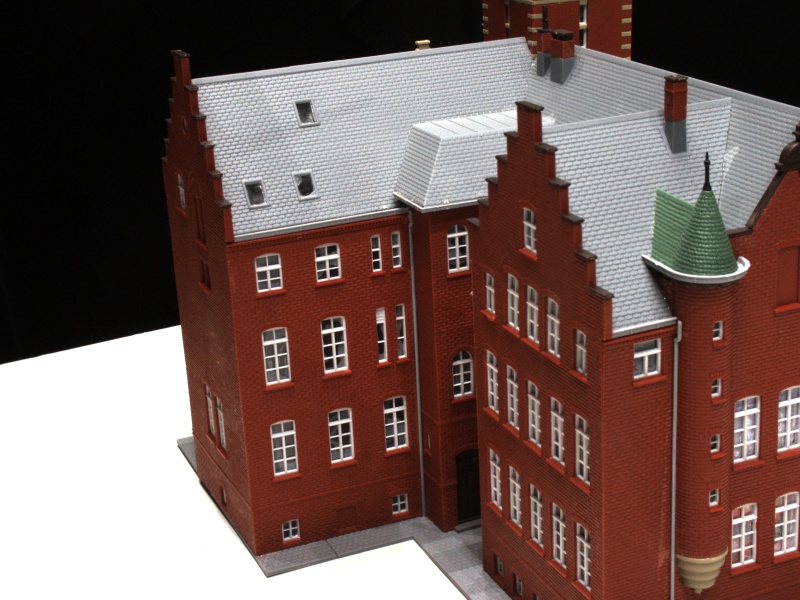}
  \end{minipage}
  ~
  \begin{minipage}[c]{0.13\linewidth}
  \includegraphics[width=0.99\linewidth]{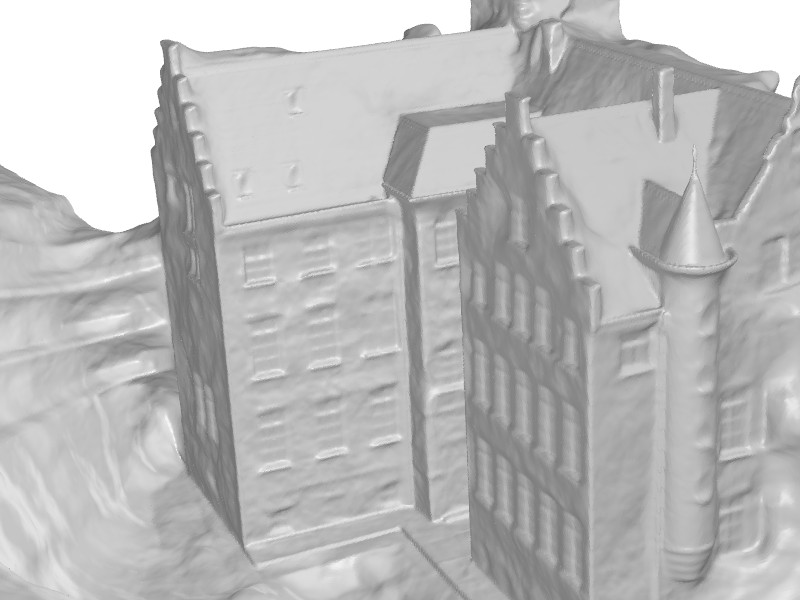}
  \end{minipage}
  ~
  \begin{minipage}[c]{0.13\linewidth}
  \includegraphics[width=0.99\linewidth]{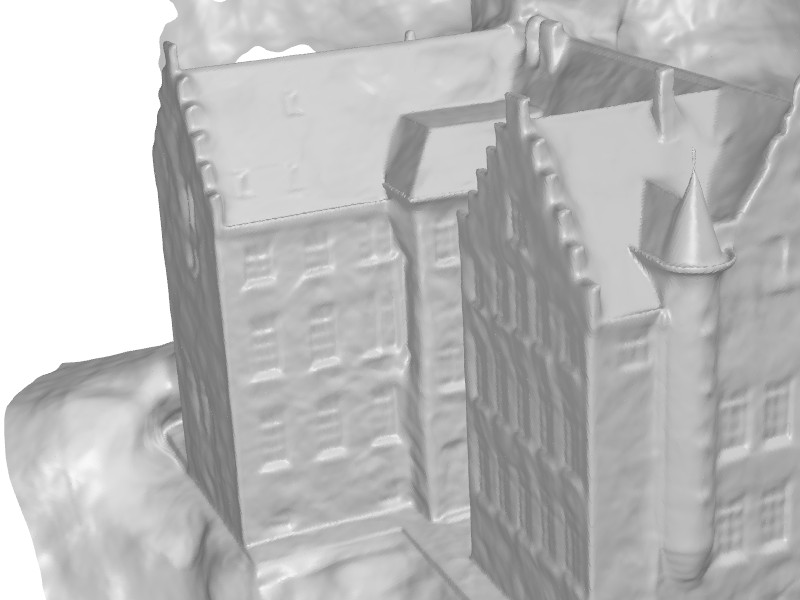}
  \end{minipage}
  ~
  \begin{minipage}[c]{0.13\linewidth}
  \includegraphics[width=0.99\linewidth]{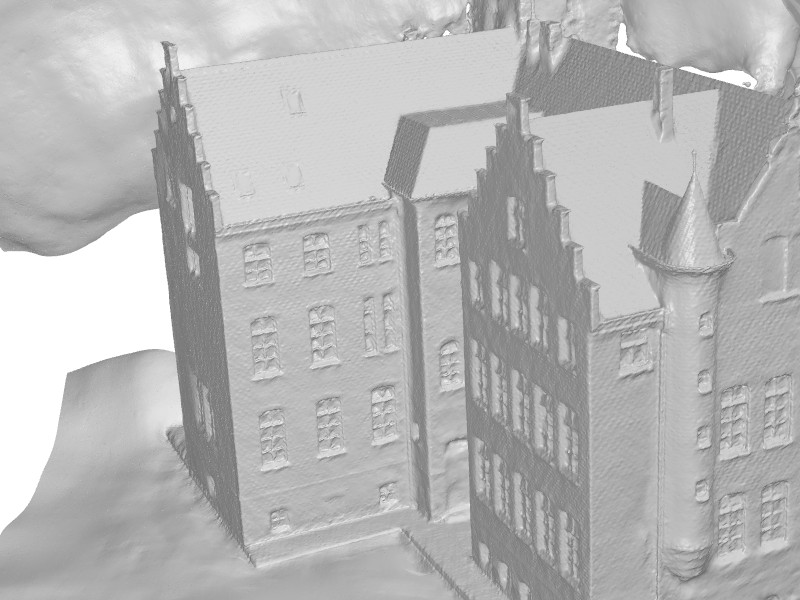}
  \end{minipage}
  ~
  \begin{minipage}[c]{0.13\linewidth}
  \includegraphics[width=0.99\linewidth]{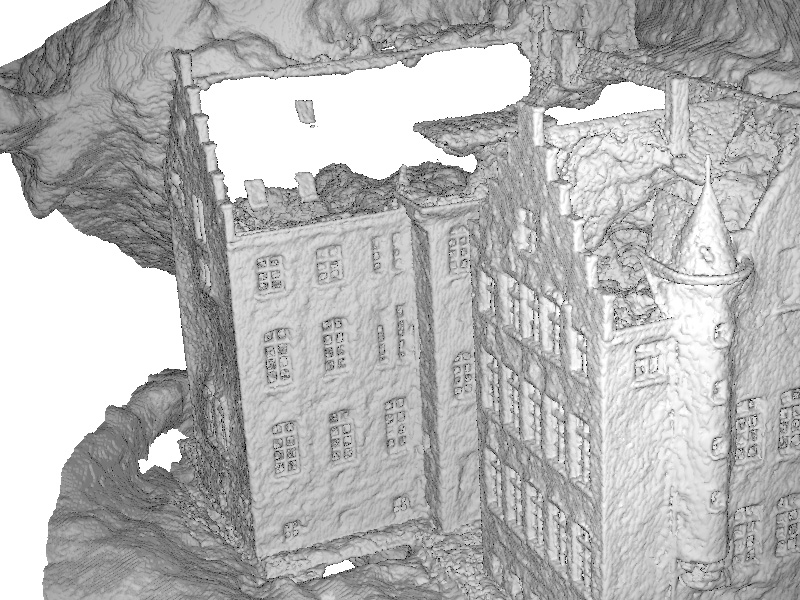}
  \end{minipage}
  ~
  \begin{minipage}[c]{0.13\linewidth}
  \includegraphics[width=0.99\linewidth]{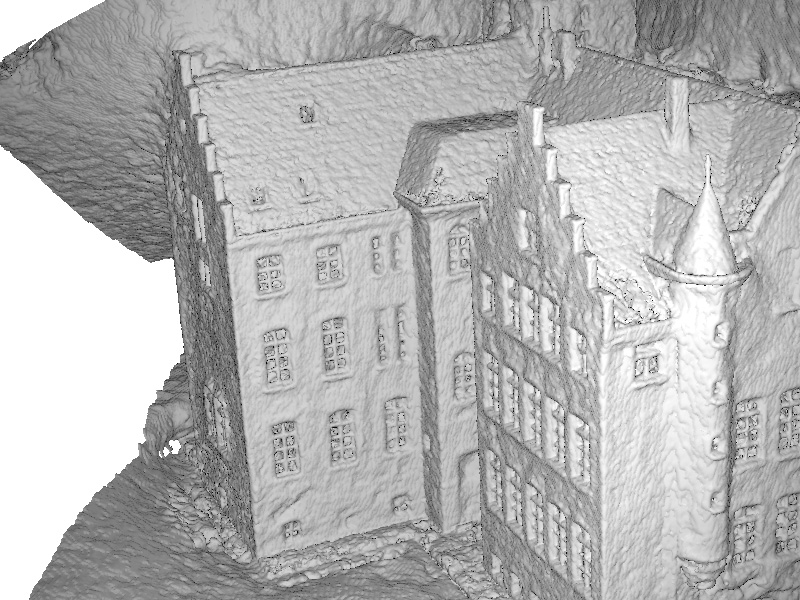}
  \end{minipage}
\end{minipage}
\\
\begin{minipage}[c]{\linewidth}
\centering
\begin{minipage}[c]{0.05\linewidth}
  \centering
  \small
  69
  \end{minipage}
  ~
  \begin{minipage}[c]{0.13\linewidth}
  \includegraphics[width=0.99\linewidth]{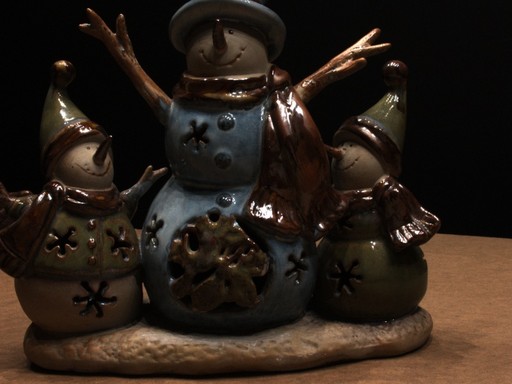}
  \end{minipage}
  ~
  \begin{minipage}[c]{0.13\linewidth}
  \includegraphics[width=0.99\linewidth]{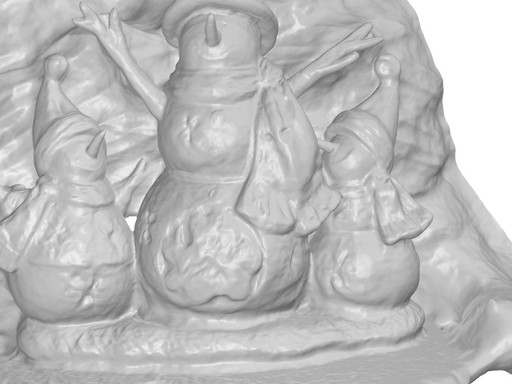}
  \end{minipage}
  ~
  \begin{minipage}[c]{0.13\linewidth}
  \includegraphics[width=0.99\linewidth]{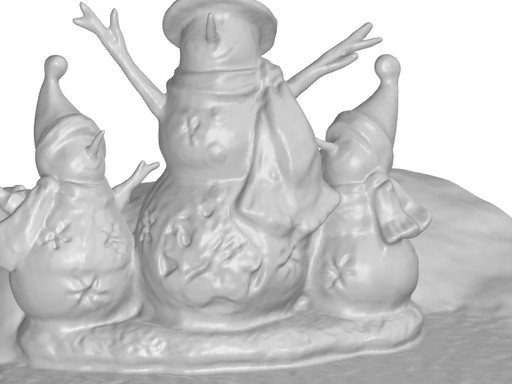}
  \end{minipage}
  ~
  \begin{minipage}[c]{0.13\linewidth}
  \includegraphics[width=0.99\linewidth]{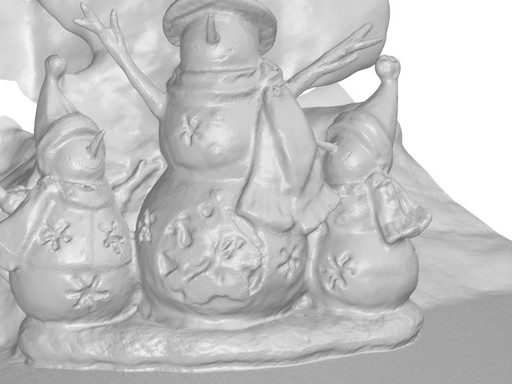}
  \end{minipage}
  ~
  \begin{minipage}[c]{0.13\linewidth}
  \includegraphics[width=0.99\linewidth]{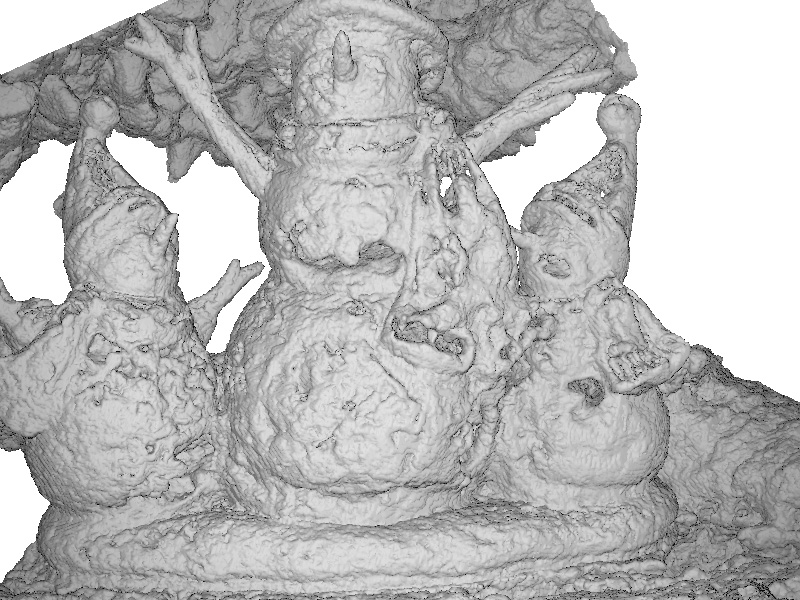}
  \end{minipage}
  ~
  \begin{minipage}[c]{0.13\linewidth}
  \includegraphics[width=0.99\linewidth]{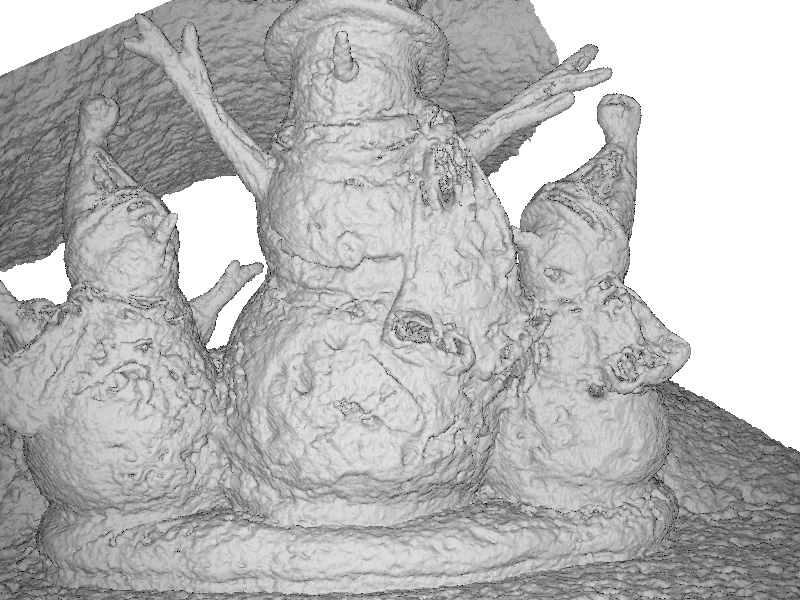}
  \end{minipage}
\end{minipage}
\\
\begin{minipage}[c]{\linewidth}
\centering
\begin{minipage}[c]{0.05\linewidth}
  \centering
  \small
  83
  \end{minipage}
  ~
  \begin{minipage}[c]{0.13\linewidth}
  \includegraphics[width=0.99\linewidth]{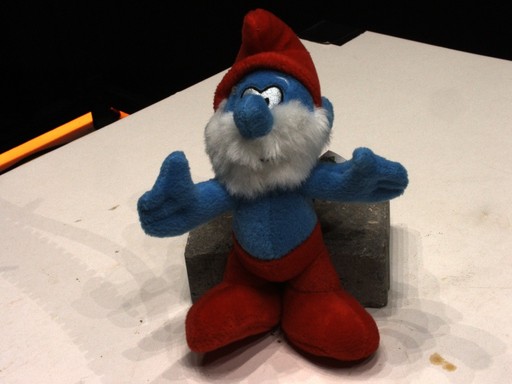}
  \end{minipage}
  ~
  \begin{minipage}[c]{0.13\linewidth}
  \includegraphics[width=0.99\linewidth]{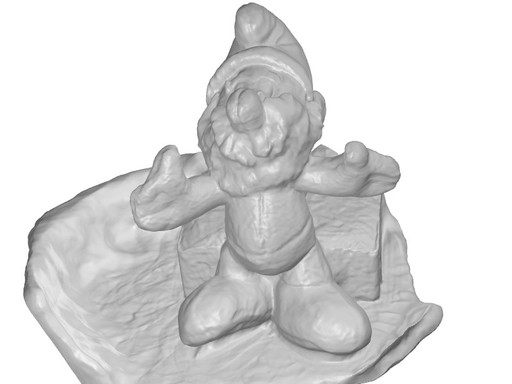}
  \end{minipage}
  ~
  \begin{minipage}[c]{0.13\linewidth}
  \includegraphics[width=0.99\linewidth]{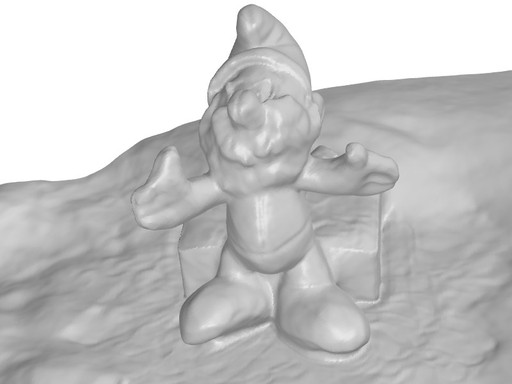}
  \end{minipage}
  ~
  \begin{minipage}[c]{0.13\linewidth}
  \includegraphics[width=0.99\linewidth]{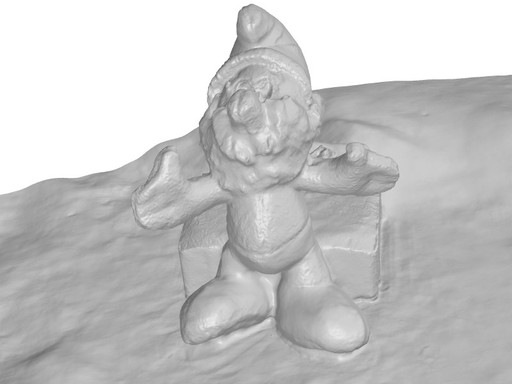}
  \end{minipage}
  ~
  \begin{minipage}[c]{0.13\linewidth}
  \includegraphics[width=0.99\linewidth]{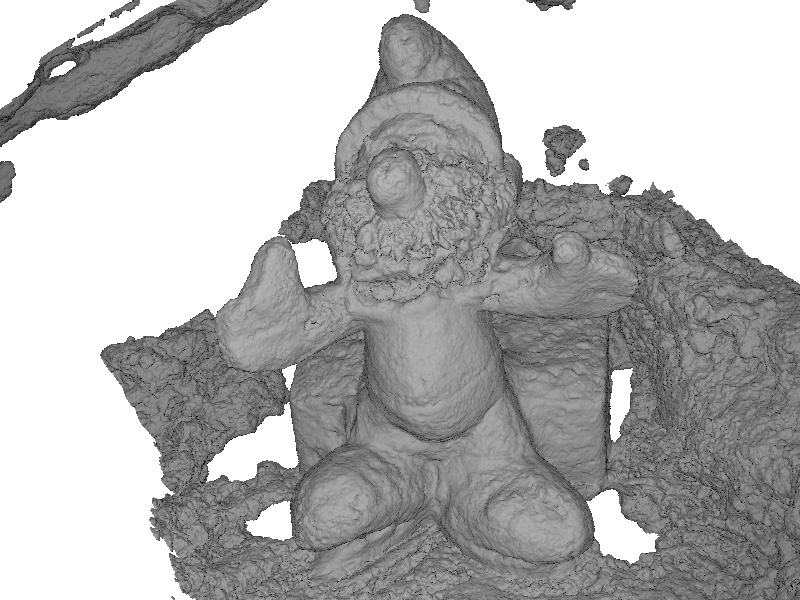}
  \end{minipage}
  ~
  \begin{minipage}[c]{0.13\linewidth}
  \includegraphics[width=0.99\linewidth]{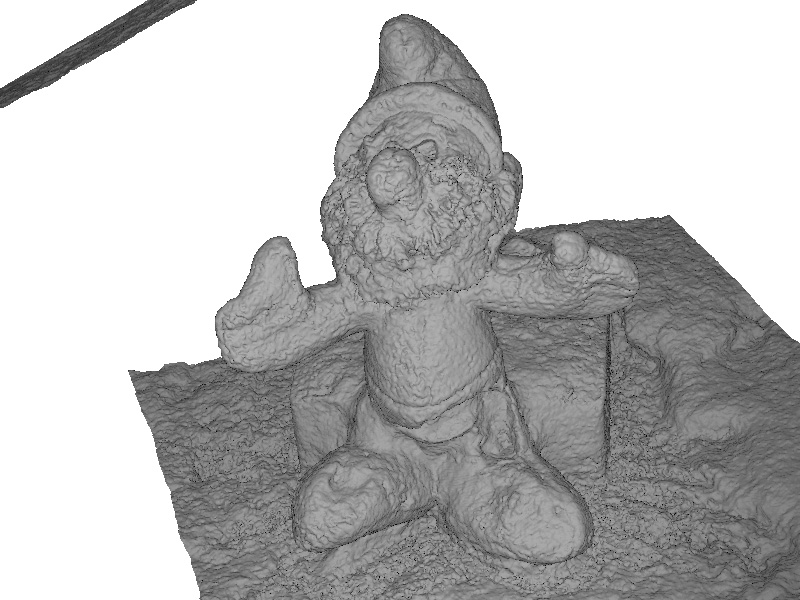}
  \end{minipage}
\end{minipage}
\\
\begin{minipage}[c]{\linewidth}
\centering
\begin{minipage}[c]{0.05\linewidth}
  \centering
  \small
  110
  \end{minipage}
  ~
  \begin{minipage}[c]{0.13\linewidth}
  \includegraphics[width=0.99\linewidth]{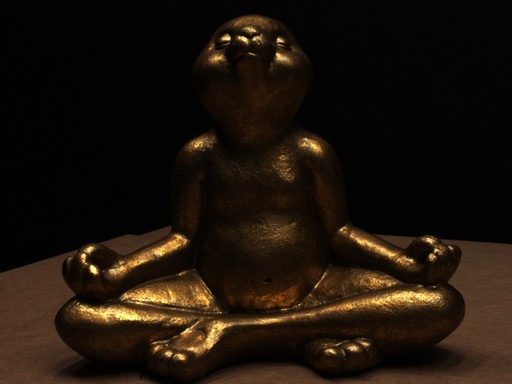}
  \end{minipage}
  ~
  \begin{minipage}[c]{0.13\linewidth}
  \includegraphics[width=0.99\linewidth]{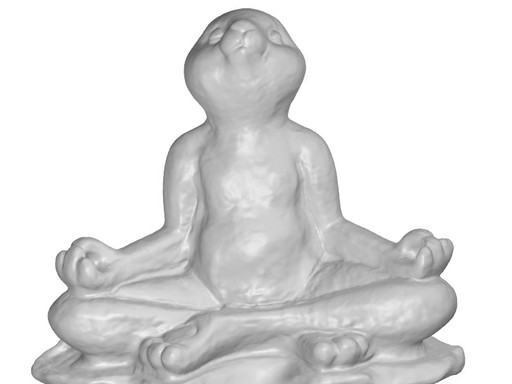}
  \end{minipage}
  ~
  \begin{minipage}[c]{0.13\linewidth}
  \includegraphics[width=0.99\linewidth]{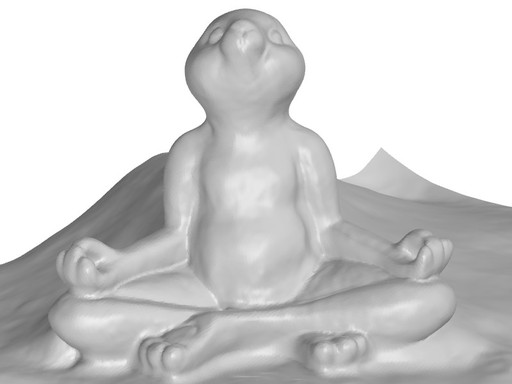}
  \end{minipage}
  ~
  \begin{minipage}[c]{0.13\linewidth}
  \includegraphics[width=0.99\linewidth]{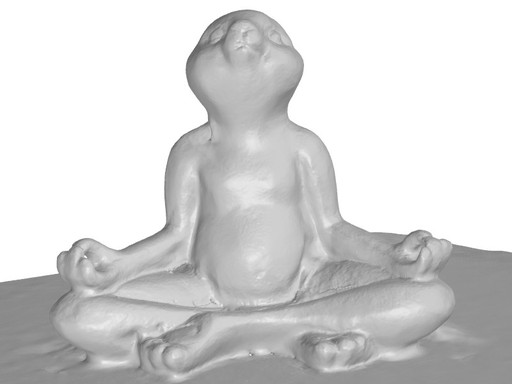}
  \end{minipage}
  ~
  \begin{minipage}[c]{0.13\linewidth}
  \includegraphics[width=0.99\linewidth]{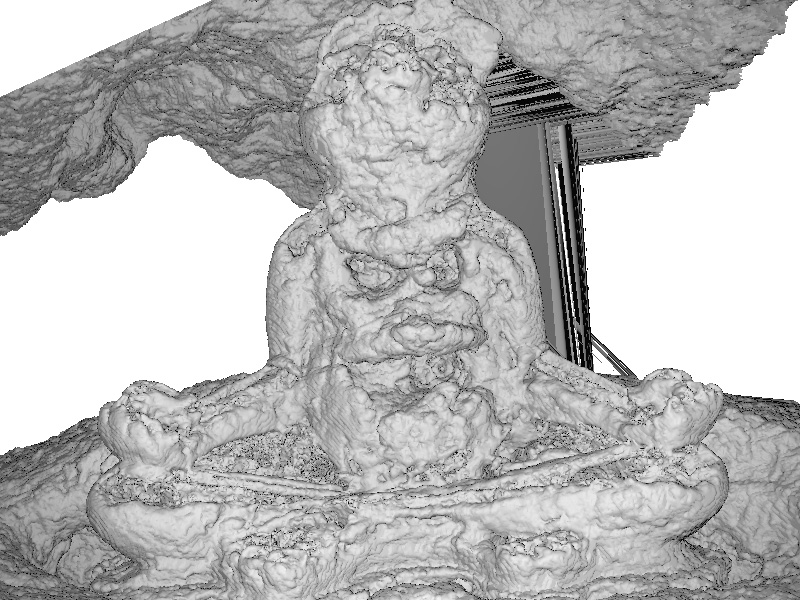}
  \end{minipage}
  ~
  \begin{minipage}[c]{0.13\linewidth}
  \includegraphics[width=0.99\linewidth]{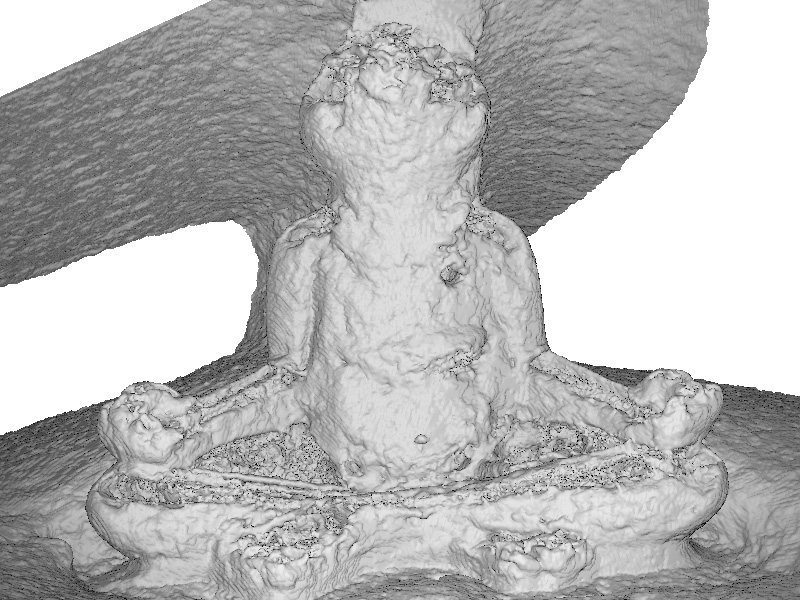}
  \end{minipage}
\end{minipage}
\end{minipage}
\\
   \caption{Qualitative comparison of our method against SOTA on geometry reconstruction evaluated on DTU dataset.
   % Top to bottom we show meshes extracted for scans 24, 69, 83 and 110.
   % Unsurprisingly, occupancy and SDF-based methods generate better meshes, while density-based methods have issues, especially with shiny objects.
   % Geometric baseline struggles with estimating complete and smooth geometry, while our DDM prior improves the reconstruction quality by removing holes and making surfaces smooth.
   }
   \label{fig:qualitative_results_dtu}
\end{figure*}

In Fig.~\ref{fig:qualitative_results_dtu} the qualitative results indicate that density based methods struggle with shiny objects (rows 2 and 4) but can have higher fidelity geometry on diffuse and textured surfaces (rows 1 and 3). The textured regions alone are not sufficient for high quality output, \eg our geometric baseline struggles to complete the geometry of a house in row 1, and our DDM model provides a complementary signal to the geometric regularizers resulting in fewer holes and smoother surfaces.

\begin{table}[b!]
  \centering
  \footnotesize
  \begin{tabular}{|l|c||l|c|}
  \hline
  SDF-based Methods
  &
  \parbox[c]{3em}{
  \centering Mean Chamfer-$L1$ $\downarrow$
  }
  &
  NeRF-based Methods
  &
  \parbox[c]{3em}{
  \centering Mean Chamfer-$L1$ $\downarrow$
  }
  \\
\hline 
\hline
\arrayrulecolor{black}
% COLMAP~\cite{schoenberger2016sfm}
% & 1.36
% &&
% \\
% \hdashline
UNISURF~\cite{Oechsle2021ICCV}
& 1.02
&
Instant NGP~\cite{mueller2022instant}
& 1.71
\\
\rowcolor{LightGray}
NeuS~\cite{wang2021neus}
& 0.84
&
NeRF~\cite{mildenhall2020nerf}
& 1.49
\\
VolSDF~\cite{yariv2021volume}
& 0.86
&
Geometric Baseline
& 1.36
\\
\rowcolor{LightGray}
MonoSDF~\cite{Yu2022MonoSDF}
& 0.73
&
\textbf{DiffusioNeRF}
& 1.21
\\
\arrayrulecolor{gray}\hline
\arrayrulecolor{black}

  \end{tabular}
  \caption{\textbf{DiffusioNeRF \vs SOTA} in geometry reconstruction on the DTU dataset with all views~\cite{monodepth}.
  }
  \label{tab:recon}
\end{table}

\subsection{Ablation studies}\label{sec:ablation}
In table~\ref{tab:ablation} we show contributions of each of our optimization terms evaluated on LLFF and DTU datasets for novel view synthesis and reconstruction quality. As reported, the geometric baseline scores favorably on the LLFF dataset, but has issues in geometry as reflected in DTU scores. Qualitative results in Fig.~\ref{fig:qualitative_results_llff} demonstrate that the geometry estimated by the geometric baseline is not realistic, even if the appearance scores are high.
Our DDM-based approach improves on DTU scores, but its performance on the novel view synthesis metrics is hampered by its tendency to introduce details in areas of the scene that are not pictured in any training view.

In table~\ref{tab:ablation} we also show ablations of some of the finer details of our model. This table suggests that a model trained on $24\times24$ patches outperforms a model trained on $48\times48$ patches on LLFF, but underperforms on DTU.

\begin{table*}[ht]
  \centering
  \footnotesize
  \begin{tabular}{|l|ccc|ccc|c|}
  \hline
  & \multicolumn{3}{c|}{LLFF}
  & \multicolumn{4}{c|}{DTU}
  \\
  \cline{2-8}
  % \rule{0pt}{0.2cm}
  {\centering Method}
  &
  \multicolumn{3}{c|}{\parbox[c]{4.3em}{\centering Average $\downarrow$}} &
  \multicolumn{3}{c|}{\parbox[c]{4.3em}{\centering Average $\downarrow$}}
  &
  % \parbox[c]{4.2em}
  {\centering Chamfer-$L1$ $\downarrow$}
  % }
  \\
  \cline{2-8}
  & {\tiny  3-view} 
  & {\tiny  6-view} 
  & {\tiny  9-view}
  & {\tiny  3-view} 
  & {\tiny  6-view} 
  & {\tiny  9-view}
  & {\tiny  All views}
  \\
\hline
% \arrayrulecolor{black}

$\nabla\mathcal{L} = \nabla\mathcal{L}_{\text{photo}}$
& 0.210 & 0.128 & 0.090
 & 0.203  & 0.142  & 0.119 
& 2.87
\\
\rowcolor{LightGray}
$\nabla\mathcal{L} = \nabla\mathcal{L}_{\text{photo}} + \lambda_\text{fg}\nabla \mathcal{L}_{\text{fg}}$
& 0.210 & 0.128 & 0.090
 & 0.195  & 0.126  & 0.092
& 1.71
\\

$\nabla\mathcal{L} = \nabla\mathcal{L}_{\text{photo}} + \lambda_\text{fg}\nabla \mathcal{L}_{\text{fg}} + \lambda_\text{fr}\nabla \mathcal{L}_{\text{fr}}$
& 0.135 & 0.089 & 0.072
 & 0.215  & 0.128  & 0.093 
& 1.71
\\

\rowcolor{LightGray}

$\nabla\mathcal{L} = \nabla\mathcal{L}_{\text{photo}} + \lambda_\text{fg}\nabla \mathcal{L}_{\text{fg}} + \lambda_\text{fr}\nabla \mathcal{L}_{\text{fr}} -\lambda_\text{DDM} \epsilon_\theta$
& 0.145 & 0.085 & 0.066
 & 0.190  & 0.097  & 0.072 
& 1.67
\\

$\nabla\mathcal{L} = \nabla\mathcal{L}_{\text{photo}} + \lambda_\text{fg}\nabla \mathcal{L}_{\text{fg}} + \lambda_\text{fr}\nabla \mathcal{L}_{\text{fr}}
+\lambda_\text{dist}\nabla \mathcal{L}_{\text{dist}}
$
& 0.118 & 0.071 & 0.060
 & 0.185  & 0.092  & 0.056 
& 1.36
\\
\rowcolor{LightGray}
$\nabla\mathcal{L} = \nabla\mathcal{L}_{\text{photo}} + \lambda_\text{fg}\nabla \mathcal{L}_{\text{fg}} + \lambda_\text{fr}\nabla \mathcal{L}_{\text{fr}}
+\lambda_\text{dist}\nabla \mathcal{L}_{\text{dist}}
-\lambda_\text{DDM} \epsilon_\theta$
& 0.127 & 0.075 & 0.064
 & 0.135  & 0.052  & 0.033 
& 1.21
\\
\hdashline

DDM regularizer using 24x24 patches
& 0.126 & 0.074 & 0.061
 & 0.195  & 0.068  & 0.043 
& 1.22
\\
\rowcolor{LightGray}
24x24 patch DDM \& NeRF fitted with $4\times \lambda_\text{DDM}$
& 0.129 & 0.074 & 0.062
 & 0.260  & 0.080  & 0.050 
& 1.22
\\
Patches from input images are not given to DDM
& 0.139 & 0.078 & 0.066
 & 0.159  & 0.063  & 0.049 
& 1.91
\\
\rowcolor{LightGray}
DDM trained with 20\% of Hypersim scenes
& 0.132 & 0.078 & 0.066
 & 0.163  & 0.057  & 0.035 
& 1.65
\\
RGB-only DDM regularizer
& 0.134 & 0.083 & 0.070
 & 0.189  & 0.081  & 0.058 
& 1.31
\\
\rowcolor{LightGray}
$\tau=0$ (no schedule) during NeRF fitting
& 0.137 & 0.081 & 0.067
 & 0.152  & 0.055  & 0.042 
& 1.31
\\
NeRF fitted with $4\times \lambda_\text{DDM}$
& 0.146 & 0.088 & 0.076
 & 0.220  & 0.134  & 0.071 
& 2.56
\\

\arrayrulecolor{gray}\hline
\arrayrulecolor{black}

  \end{tabular}
  \caption{Ablation study of our method. Note that for DTU, $\lambda_\text{fr}$ is set to 0, hence the 2nd and 3rd rows have identical scores on DTU. Geometric baseline corresponds to the model in the 5th row. Note that for these ablations we use LPIPS-Alex, rather than LPIPS-VGG; see section 5 of the supplemental for more details.}
  \label{tab:ablation}
\end{table*}

The ablations show the significance of feeding patches from input images to DDM 25\% of the time during NeRF fitting. It can be especially important early on, when rendered patches are very different from input images.

Unsurprisingly, reducing the amount of training data for the DDM (only using 20\% of the Hypersim scenes) slightly reduces the scores. The RGB-only regularization with DDMs is similar to RegNeRF's normalizing flow model regularization, but with larger patch sizes. Interestingly, the RGBD regularizer trained with 20\% of the data is still better than the RGB-only regularizer that was trained with 100\% of the data. The last two rows of the ablation show that careful scheduling of $\tau$ and DDM gradient weights is necessary to produce good results. This is an active area of research, having previously been noted in~\cite{poole2022dreamfusion}. The DDM weight $\lambda_\text{DDM}$ trades off the accuracy of reconstruction around thin structures against the overall depth smoothness.

%------------------------------------------------------------------------

\section{Conclusions}\label{sec:conclusions}
In this paper we address the problem of regularization of NeRFs. Our approach uses a DDM trained on RGBD patches to approximate a score function, \ie the gradient of the logarithm of an RGBD patch distribution. Experimentally, we demonstrate that the proposed regularization scheme improves performance on novel view synthesis and 3D reconstruction.

While we show regularization using color and depth patches as input, the proposed framework is versatile and can be used to regularize the 3D voxel grid of densities, density weights sampled along the ray, \etc. Indeed, instead of generating RGBD patches, we can generate 3D voxel blocks of densities to train a DDM and use it during NeRF optimization to regularize the density field directly. Early results are shown in the supplementary materials.

% One avenue of future work is formulating a principled approach of combining the DDM gradient with NeRF objective to avoid heuristics-based $\tau$ and loss term weight scheduling. 
One avenue of future work is formulating a principled combination of the DDM gradient with the NeRF objective to avoid heuristics-based $\tau$ and gradient scheduling. 

Our work is focused on NeRF optimization, however the general approach of using DDMs as a regularizer could potentially be used for other tasks that are optimized with gradient descent, \eg self-supervised monocular depth estimation~\cite{monodepth}, or self-supervised stereo matching~\cite{zhong2017self,zhou2017unsupervised}.

\newpage
\noindent
\textbf{Acknowledgements}
We thank Niantic colleagues, especially Gabriel Brostow, for discussions and suggestions. We are also grateful for Jiaxiang Tang's Pytorch implementation of Instant-NGP~\cite{torch-ngp}, Phil Wang's implementation of DDM~\cite{torch-ddm}, and to Thomas M\"uller for tiny-cuda-nn~\cite{tiny-cuda-nn}. We also thank Jones Smith for bringing to our attention an error with our use of LPIPS in a prior version of this paper.

%%%%%%%%% REFERENCES
{\small
\bibliographystyle{ieee_fullname}
\bibliography{egbib}
}

\end{document}